\pdfoutput=1

\documentclass[11pt]{article}

\usepackage{emnlp2021}

\usepackage{times}
\usepackage{latexsym}

\usepackage{siunitx}
\usepackage[T1]{fontenc}

\usepackage[utf8]{inputenc}

\usepackage{todonotes}
\usepackage{subcaption}

\usepackage{multirow}
\usepackage{caption}
\usepackage{amsmath,amssymb}
\usepackage{array}
\usepackage{booktabs, longtable, supertabular}
\usepackage{graphicx}

\usepackage{rotating}

\usepackage{tikz,graphics,color,float,epsf}
\usepackage{hyperref}
\usepackage{cleveref}
\usepackage{url}

\usepackage{lscape}

\usepackage{microtype}

\title{Beyond Distillation: Task-level Mixture-of-Experts for Efficient Inference}

\author{Sneha Kudugunta, Yanping Huang, Ankur Bapna, Maxim Krikun, \\
\textbf{Dmitry Lepikhin, Thang Luong, Orhan Firat} \\
Google Research \\
\texttt{\{snehark,huangyp,ankurbpn,krikun,lepikhin,} \\
\texttt{thangluong,orhanf\}@google.com} \\
}

\begin{document}
\maketitle
\begin{abstract}
Sparse Mixture-of-Experts (MoE) has been a successful approach for scaling multilingual translation models to billions of parameters without a proportional increase in training computation. However, MoE models are prohibitively large
and practitioners often resort to methods such as 
distillation for serving. 
In this work, we investigate routing strategies at different granularity (token, sentence, task) in MoE models to bypass distillation. 
Experiments on WMT and 
a
web-scale dataset suggest that task-level routing ({\it task-MoE}) enables us to extract smaller, ready-to-deploy sub-networks from large sparse models.

On WMT, our task-MoE with 32 experts (533M parameters) outperforms the best performing token-level MoE model ({\it token-MoE}) by +1.0 BLEU on average across 30 language pairs. 
The peak inference throughput is also improved by a factor of 1.9x when we route by tasks instead of tokens.
While distilling a token-MoE to a smaller dense model preserves only $32\%$ of the BLEU gains, our sub-network task-MoE, by design, preserves all the gains with the same inference cost as the distilled student model. 
Finally, when scaling up to 200 language pairs, our 128-expert task-MoE (13B parameters) performs competitively with a token-level counterpart, while improving the peak inference throughput by a factor of 2.6x.
\end{abstract}

\section{Introduction}\label{sec:intro}

Scaling up neural network models has recently received great attention, given the significant quality improvements on a variety of tasks including natural language understanding  \citep{raffel2019exploring,brown2020language} and multilingual machine translation \citep{huang2019gpipe,lepikhin2020gshard}.

While training massive models on large amounts of data can almost guarantee improved quality, there are two factors affecting their practicality and applicability: (1) \textit{training efficiency} and (2) \textit{inference efficiency}. Large dense models are often prohibitively compute-intensive to train, with some models requiring TFlops-days of compute \citep{brown2020language}. A recent line of work has proposed sparsely-gated Mixture-of-Experts (MoE) layers as an efficient alternative to dense models \citep{shazeer2017outrageously,lepikhin2020gshard,riabinin2020learning} in order to address \textit{training efficiency} limitations. In a vanilla sparsely-gated MoE model each token of the input sequence activates a different subset of the experts, hence the computation cost per token becomes only proportional to the size of the activated sub-network. However, they fail to meet requirements on \textit{inference efficiency}. 

Consider a long sequence where each token of the sequence activates a disjoint subset of available experts. From a practical standpoint, the inference trace of the full sequence spans several experts independently for every token, resulting in an independent pathway for each token. Although this is a desired property - adding flexibility to the model and increasing its capacity - it becomes prohibitive for inference for the following reasons: the model parameters in these large models are beyond the memory limit of a single accelerator device, and require model parallelism to shard them across a cluster of devices during inference. For models with MoE Layers, the input token would be dynamically routed to different experts allocated to different devices. This further adds communication cost across devices to the overall serving cost. Moreover, due to the sequential nature of the auto-regressive decoding~\citep{kasai2020deep, chen-EtAl:2018:Long1}, the added communication cost from model parallel decoders gets multiplied by the number of decoding steps.  To add to this, serving MoE models efficiently requires batching a large number of input tokens together, otherwise only a subset of the MoE network will be activated leading to severe device under-utilization. 

In this work, we study the \textit{inference efficiency} of sparsely gated MoE models while taking into account the characteristics of the intended application, Multilingual Neural Machine Translation (MNMT). MNMT is an inherently multi-task learning problem, aimed at building a single neural network for translating multiple language pairs simultaneously. In a MNMT model, the extent to which parameters are shared across languages determines the magnitude of positive transfer \citep{baldwin1988transfer} and conversely task interference due to the capacity bottleneck \citep{arivazhagan2019massively}. In an ideal scenario, we would want to efficiently train a single large MNMT model maximizing transfer while expanding the capacity bottleneck; meanwhile, we would like to enjoy the benefits of sparsely activated sub-networks per-task at inference time, i.e. extracting out a sub-network to decode for a particular language pair to actualize \textit{inference efficiency}. 

An alternative way to enjoy high inference efficiency from a large model is knowledge distillation~\citep{hinton2015distilling}. However, ~\citep{DBLP:journals/corr/abs-2101-03961} found that only a small fraction of quality gains  from a large sparse model can be preserved in the student models. Instead;

\begin{itemize}
\item We propose routing algorithms for MoE models with affordable serving costs (Section~\ref{sec:methods}). While vanilla MoEs route each sub-word token in the input to its preferred experts, we explore alternative routing strategies that are trained to leverage global task level information to route all tokens corresponding to a particular task collectively to the same set of experts. We decode different tasks separately and only load the subset of experts associated with the corresponding task during inference. 


\item We report the advantages of our task-level routing method in translation quality and inference cost on a multilingual WMT task (Section~\ref{sec:exp}). With the comparable inference cost, the task-level routing achieved +$3.6$ BLEU gain over the multilingual model training from scratch, and +$2.1$ BLEU gain over the dense student model distilled from the large token-level /position-wise MoE (token-MoE) model. 
\item The observed quality gains from our approach are comparable with the token-MoE models while achieving 1.9x  peak throughput and $6.3$\% of the decoder size. 
\item We scaled up the token-MoE model on a large scale in-house dataset and saw similar quality gains (+$3.6$ BLEU) against the dense baseline (Section \ref{sec:m4-res2}). Compared to the token-level routing approach, our method achieves comparable quality gain, with $2.6$x higher peak throughput and $1.6$\% of the decoder size. 
\item Finally, we analyze the routing decisions made in MoE models and motivate our method (Section \ref{sec:moe-analysis}). 

\end{itemize}

\section{Scaling Transformers with Mixture-of-Experts}\label{sec:xf-moe}

\newcommand{\combine}{\mathcal{G}}  
\newcommand{\gates}{g}    

The Transformer \citep{vaswani2017attention} architecture is a popular model used for neural machine translation and other natural language understanding/generation problems. In sequence-to-sequence problems, the model consists of an encoder and decoder, each of which contains multiple Transformer layers. For further details, we refer the reader to the original paper \citep{vaswani2017attention}.

We use the Mixture-of-Experts Transformer models proposed by \cite{lepikhin2020gshard}, where the MoE layers for the Transformers consist of $E$ feed-forward networks (FFN), such that ($\text{FFN}_1 \dots \text{FFN}_E$).

\vspace{-0.2in}
\begin{align*}\small
    \text{FFN}_e(x_s)         & = wo_e \cdot \text{ReLU}(wi_e \cdot x_s)                     \\
    y_s                    & = \sum^E_{e=1} \combine_{s,e} \cdot \text{FFN}_e(x_s)      
\end{align*}
Here,  $x_s$ is the input token at position $s$ to the MoE layer and each $\text{FFN}_e$ is a two layer neural network using a ReLU activation function. $wi_e$ and $wo_e$ are the input and output projection weights of the $e$-th expert. Finally, $\combine_{s,E}$ is vector computed by the gating network (also referred as router). For each expert, most values of this vector are zeros, one value being positive. We use this vector to route the token to a select few experts. The entries chosen from  $\combine_{s,E}$ determine how much the expert contributes to the final output $y_s$. Note that, in this work we choose the top 2 weight experts for each example to be comparable with the prior work.

\begin{figure*}

\begin{subfigure}{0.55\textwidth}
    \includegraphics[width=1.0\textwidth]{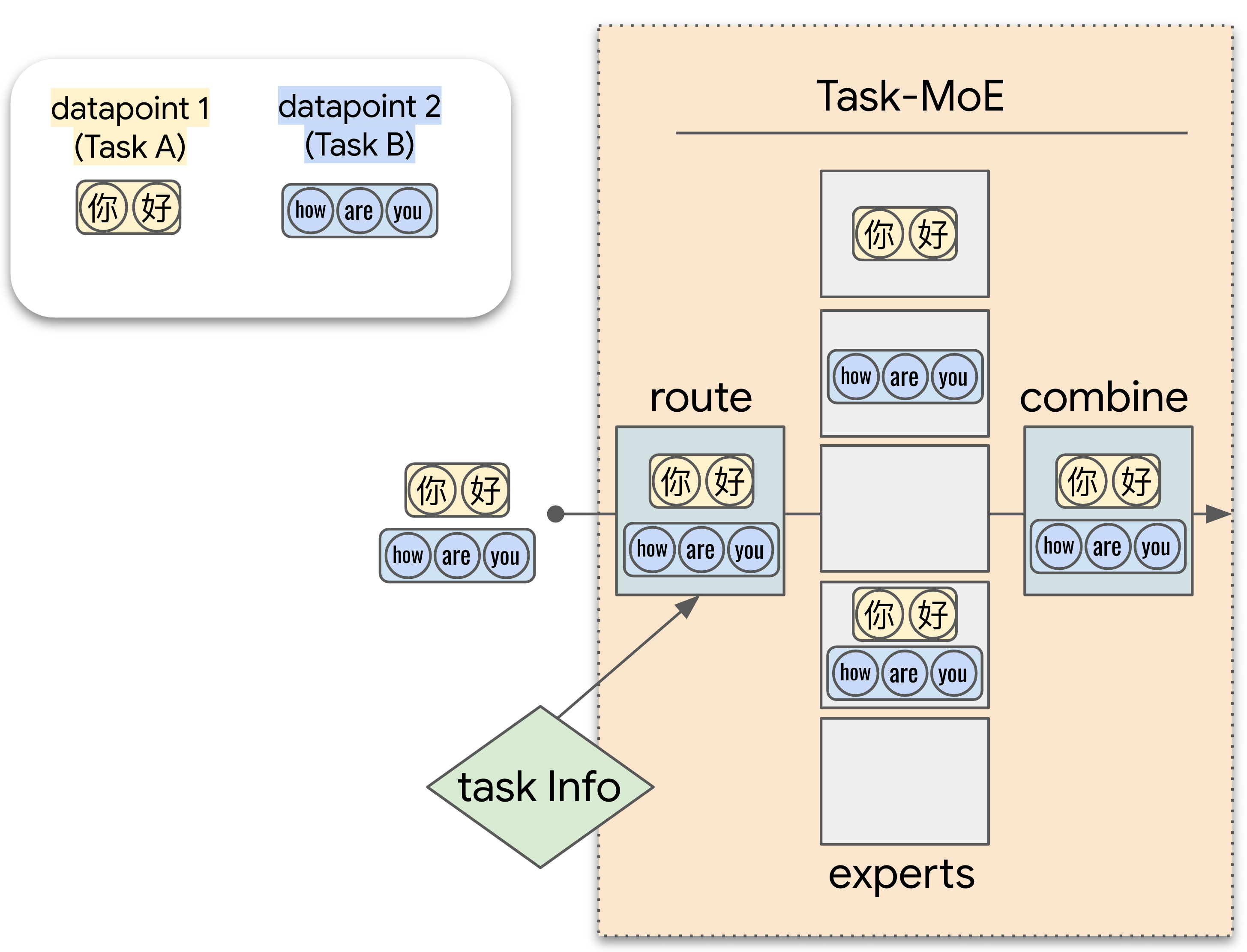}
    \caption{Task MoE}    
    \label{fig:task-moe}
\end{subfigure}
\begin{subfigure}{0.45\textwidth}
    \centering
    \includegraphics[width=0.8\textwidth]{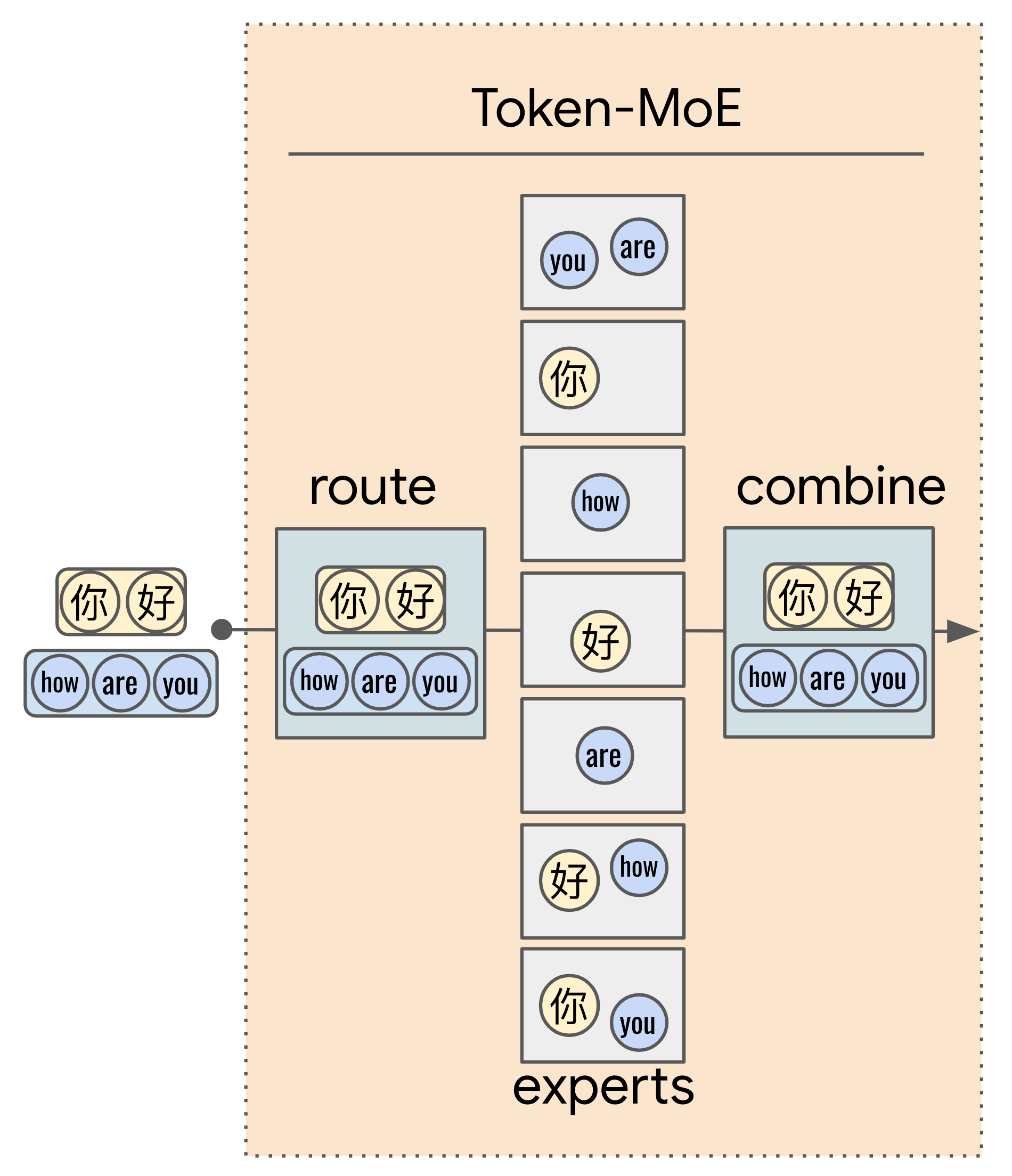}
    \caption{Token MoE}
    \label{fig:moe}
\end{subfigure}

\caption{Tokens are routed to the same expert based on task or some other prior in (a) task-based MoE whereas different tokens are routed to different experts in (b) token-based MoE models.}
\label{fig:moe-types}
\end{figure*}

The gating network $\combine_{s,E}$ must be considered carefully for efficiency purposes: (1) the utilization of experts must be balanced and (2) the function must be efficient to implement at scale. For a more thorough discussion of MoE transformers, we direct the reader to \cite{lepikhin2020gshard}.

\section{Methods}\label{sec:methods}


In this section we describe our candidate routing strategies in the context of MNMT and discuss their trade-offs from the perspective of the training and inference efficiency. Multilingual models learn joint representations across languages to the extent of the parameters being shared \citep{wu2019beto,tiedemann2018emerging,tan2019multilingual,zhang2020improving,ostling2016continuous,kudugunta2019investigating}. While being beneficial for transfer, extreme sharing of the parameters exacerbates interference. Allowing dedicated (unshared) parameters are known to be effective at mitigating interference \cite{zhang2021share,kong2021multilingual} and MoE variants are inherently learn such partitioning across languages/tasks. Therefore we study the routing algorithm $\text{GATE}(x_s)$ of MoEs to mitigate interference, while enabling transfer and effective at inference.

\subsection{Routing Strategies}\label{sec:routing}

Given the sequential nature of the multilingual machine translation task, the routing decisions can be made at three different granularities, from bottom up (i) token-level, (ii) sentence-level and (iii) task-level, as detailed below.

\paragraph{Token-level Routing:} This is the baseline discussed in Section \ref{sec:xf-moe} where each token is routed independently.
\paragraph{Sentence-level Routing:} Each sequence (sentence), and all tokens that form the sequence, are routed to the same expert. We change the routing algorithm to select experts by sentence representation, calculated by taking the average token representations in a given sentence.
\paragraph{Task-level Routing:} We  select experts by task boundaries as opposed to making input-level decisions. In the context of MNMT, these task boundaries can either be defined by the target language (French-to-English and German-to-English are the same task) or the language pair (French-to-English and German-to-English are different tasks). Sentence and task level routing are formulated as follows:

\vspace{-5px}

\begin{align*}
    \combine_{s,E}           & = \text{GATE}(\frac{1}{S}\sum^S_{s=1} x_s )   \texttt{   }  \text{ (Sentence-level),}            \\
    \combine_{s,E}           & = \text{GATE}(\text{task\_id}_s)      \texttt{    } \text{ (Task-level).}  
\end{align*}

We illustrate the difference in Figure \ref{fig:moe-types}, in token-based MoE models (Figure \ref{fig:moe}), tokens from each datapoint are routed to different experts, whereas in task-level MoE models (Figure \ref{fig:task-moe}), tokens may be routed to the same expert based on task.


\subsection{Inference Implications of Routing Strategies}\label{sec:tradeoffs}

While the MoE models discussed in \citep{shazeer2017outrageously,lepikhin2020gshard} train quickly relative to the number of parameters in terms of the wall-clock time, they are expensive to serve. Consider a MoE with 512 experts and 50B parameters~\citep{lepikhin2020gshard}. When employing token-level routing, each token can be independently routed to a different set of experts during inference. Given that the entire model is too large to load into memory on a single accelerator, the two potential solutions to utilize this model for inference are: (i) Loading experts dynamically from host to device depending on routing decisions, or (ii) Utilizing model-parallelism over multiple accelerators for serving. While the first solution incurs heavy host-device communication costs, the second introduces  significantly inter-device communication overhead.

Other practical approaches to serve a large MoE include model quantization, pruning and knowledge distillation~\citep{cheng2017survey}. While the first two strategies haven't been explored in the context of  conditional computation, distillation \citep{hinton2015distilling,kim-rush-2016-sequence} has been found to introduce undesirable artifacts into the student model~\citep{freitag-etal-2019-ape,bogoychev2019domain} in the context of NMT. Moreover, some studies have found that distilling large sparse models preserves only a small fraction of the gains achieved by scaling. On the other hand, if we limit the number of experts available to every task in the model to a small fraction of the total available capacity, it is possible to extract task-specific models for serving, alleviating the need for complex serving strategies or compression. Since decoding time complexity for auto-regressive encoder-decoder models is dominated by the decoder~\citep{kasai2020deep}, we can also pursue a hybrid strategy where the encoder utilizes more expensive routing strategies while the decoder of the model utilizes simpler and efficient routing.


Summarizing the \textit{effective} decoding cost of the MoE models utilizing different routing strategies:
\begin{itemize}
\item \textbf{Token/Sentence level routing}: The routing decisions are made dynamically. Assuming each token/sentence makes disjoint choices, the server needs to load all $E$ experts.
\item \textbf{Task-level routing}:  Tokens corresponding to each input sentence are routed to the same experts statically. The server only needs to pre-load $K$ experts (assuming top-$K$ routing).
\end{itemize}

\section{Experiments on 30 Language Pairs}\label{sec:exp}

We compare routing strategies at multiple levels in both, the encoder and the decoder, by conducting extensive experiments on two benchmarks: the public WMT dataset with 30 language pairs (Section \ref{sec:wmt-setup}) and an in-house web-scale dataset with 200 language pairs (Section \ref{sec:m4-exp}). We start with WMT setup.

\subsection{Experimental Setup}\label{sec:wmt-setup}

For our experiments, we use parallel training and evaluation data from the WMT corpus and adopt the setup used by \cite{siddhant2020leveraging} with 15 languages, to and from English. Full training data details may be found in Table \ref{tab:dataset-bi} in the Appendix. The amount of data ranges from more than 60 million sentence pairs in en-cs translation direction (en-cs) to roughly 150k sentence pairs for en-gu. 

We use a temperature based data sampling strategy to train our models, similar to the strategy used to train the multilingual models in \cite{arivazhagan2019massively}: if $p_L$ is the probability that a sentence in the corpus belongs to language pair $L$, we sample from a distribution where the probability of sampling from $L$ is proportional to ${p_L}^{\frac{1}{T}}$. All the experiments in this paper are performed on a model trained with a sampling temperature $T=5$.

We use the 142M Transformer Base \citep{vaswani2017attention} architecture (or enhanced versions of it with MoE layers) for all of our experiments with WMT. Our models are optimized using Adafactor \citep{shazeer2018adafactor} with momentum factorization and a per-parameter norm clipping threshold of 1.0. We followed a learning rate of 3.0,  with 40K warm-up steps for the schedule, which is decayed with the inverse square root of the number of training steps after warm-up. BLEU scores presented in this paper are calculated using SacreBLEU \cite{post-2018-call} on the WMT test sets.

\begin{table*}[t!]
\resizebox{\linewidth}{!}{

\begin{tabular}{c|c|c|c|c|cc|cc}
\toprule
\multirow{2}{*}{{\bf System}}& \multicolumn{2}{c|}{{\bf Routing Granularity}} & {\bf Throughput}  &\multicolumn{5}{c}{{\bf BLEU}} \\
\cline{2-9}
& Encoder & Decoder & Peak tokens/s &\bf{Average} & xx2en & en2xx & High  & Low \\
\hline
Bilingual Baselines  &    -   &     -   & \multirow{2}{*}{\num{2.3e5} }  &  21.0 & 21.8 & 18.9 & 28.2 & 11.8 \\
Multilingual Transformer-Base   &    -   &     -   &  &  20.0 & 23.7 & 17.5 & 23.3 & 15.9 \\
\hline
Static MoE -- 32 experts     &   -    &  -   & \num{2.3e5} & 17.6 & 25.0 & 10.2 & 20.9 & 13.5 \\
\hline
Token-level MoE -- 32 experts     &   Token    &     Token   & \num{1.3e5} & 22.6 & 24.9 & 20.4 & 27.5 & 16.3 \\

\hline
Sentence-level MoE -- 32 expert      &   Sentence    & Sentence & \num{1.3e5}  & 19.9 & 24.1 & 16.8 & 22.6 & 16.1 \\

\hline
\multirow{6}{*}{Task-level MoE -- 32 experts} &    Language Pair  &     Language Pair     & \multirow{6}{*}{\num{2.3e5}}  & 21.4 & 25.2 & 16.9 & 23.4 & 17.3   \\
 &   Target    &    Target     &  & 22.9 & 25.6 & 20.2 & 27.2 & 17.3  \\
 &    Language Pair   &      Token   &   & 22.4 & 25.6 & 20.3 & 26.9  & 16.8 \\
 &    Target   &      Token  &   &  22.3  &	24.5  &	 20.4  &	26.8 &	16.6 \\
 &  Token  &  Language Pair &    & 23.0 & \textbf{26.2} & 20.3 & 27.2 & \textbf{17.6}   \\
 &  Token     &    Target    &   & \textbf{23.6} & 26.0 & \textbf{21.1} & \textbf{28.5}  & 17.4 \\
\bottomrule
\end{tabular}
}
\caption{{\bf Routing strategies for Mixture-of-Experts (MoE) models} -- We compare routing experts by either tokens, sentence representations, or tasks (using either language pairs or target languages). For task-level MoE, routing can also be different between encoder and decoder. For results, {\it Average} is the average results of all language pairs, whereas {\it xx2en} and {\it en2xx} are the averages of translations into and from English respectively. {\it High} indicates high-resource language pairs ($>1$ million sentence pairs) while {\it Low} is for low-resource language pairs ($<1$ million sentence pairs).} 
\label{tab:32exp-compare}
\vspace{-0.1in}
\end{table*}

\paragraph{Multilingual baseline:} We train a Transformer Base model on this dataset as our multilingual dense baseline. We share all parameters across language pairs, including the softmax layer and input/output word embeddings. We use a 64k token Sentence Piece vocabulary \citep{kudo2018sentencepiece}. The vocabulary is shared on both the encoder and decoder side. Each sentence pair has a \verb <2xx>  token  pre-pended to the source sentence to indicate the target language, following~\citet{johnson2017google}.

\paragraph{Mixture of Experts Models:} For MoE models, we replace the feed forward network (FFN) of alternate layers of the Transformer with a set of identical FFN experts as depicted in Figure \ref{fig:moe}. For brevity, we provide aggregate BLEU scores in Section \ref{sec:routing-results} . We provide the full individual BLEU scores in the Appendix \ref{sec:wmt-full-res}, along with bilingual baselines. In addition, we provide the number of parameters for different components of our models in Appendix \ref{sec:param-count-wmt}.


\begin{table*}[t!]
\resizebox{\linewidth}{!}{
\begin{tabular}{c|c|c|c|c|c|c|c|c|c|c|c|c}
\toprule
\multirow{2}{*}{{\bf System}}& \multicolumn{2}{c|}{{\bf Routing Granularity}} & {\bf Throughput}  &\multicolumn{9}{c}{{\bf BLEU}} \\
\cline{2-13}
& Encoder & Decoder & Peak tokens/s & \textbf{Average} & EnFr & FrEn & EnDe & DeEn & EnRo & RoEn & EnHi & HiEn \\
\hline
Bilingual Baselines  &    -   &     -   & \num{2.3e5}  & 24.3 & 38.1 & 35.5 & 26.4 & 27.4 & 23.7 & 30.1 & 4.5 & 8.5 \\
Multilingual Transformer-Base   &    -   &     -   & \num{2.3e5}  & 25.9 & 36.1 & 34.1 & 22.0 & 28.6 & 23.9 & 33.4 & 10.4 & 19.2 \\
\hline
Task-level MoE -- 32 experts     &   Token    &     Target  & \num{2.3e5} & \textbf{29.0} & 39.9 & \textbf{37.1} & \textbf{27.1} & \textbf{32.0} & \textbf{26.6} & \textbf{36.2} & 13.3 & \textbf{20.1} \\

\hline

Token-level MoE -- 32 experts     &   Token    &     Token   & \num{1.3e5} & 28.2 & \textbf{40.1} & 36.4 & 26.7 & 31.2 & 26.5 & 33.7 & 11.5 & 19.8 \\
Distillation  (from Token MoE)   &   -    &     -   & \num{2.3e5}  & 26.9 & 37.3 & 33.2 & 25.1 & 29.3 & 24.6 & 34.6 & \textbf{13.9} & 17.6 \\
\bottomrule
\end{tabular}
}
\caption{{\bf Comparing Distillation to Task-MoE:} We compare our best performing Task-MoE model to Distilling a Token MoE model to Transformer-Base and a version with 2x the width for several language pairs. We see that distillation consistently underperforms our best-performing Task MoE model - distillation from Token MoE achieves an average BLEU score of 26.9, while our best-performing Task MoE model has an average BLEU score of 29.0 (+2.1 BLEU) for these language pairs. }
\label{tab:distill-compare}
\vspace{-0.1in}
\end{table*}


\subsection{Comparison of different Routing Strategies on WMT}\label{sec:routing-results}

We compare the token-level, sentence-level and task-level routing strategies discussed in Section \ref{sec:methods} at identical network size (32 experts, 533M parameters). The results are presented in Table \ref{tab:32exp-compare}. In general, we find that all types of task-level routing perform better than token-level routing.  We see that using sentence representations to route examples (Sentence-level MoE - 32 experts) performs much worse, so we do not conduct further experiments on this setting. In addition, we trained an MoE baseline where the experts are deterministically allocated to tasks (Static MoE - 32 Experts) - this too, did not perform well in our experiments.

When we use Task MoE on both the encoder and the decoder (Task-level MoE - 32 experts: Target/Target), we see consistent gains across the board. To investigate this further, we trained a model that has (a) Token MoE on the encoder and Task MoE on the decoder (Task-level MoE - 32 experts: Token/Target or Token/Language Pair) and (b) Task MoE on the encoder and Token MoE on the decoder (Task-level MoE - 32 experts: Target/Token or Language Pair/Token). In Table \ref{tab:32exp-compare} we see that using strategy (a) works the best, whether we choose to route by the target language or the language pair. In Section \ref{sec:moe-analysis}, we discuss these observations further.

Overall we find that using Task MoE only on the decoder (Task-level MoE 32 experts: Token/Target) works the best, with gains of 1 BLEU over Token MoE. These gains are consistent across xx2en language pairs, en2xx language pairs, high resource languages (more than 1 million sentence pairs), low resource languages and the 2 zero shot pairs.

\subsection{Comparison of Throughput of Sparse Models}\label{sec:wmt-throughput}

\begin{figure}[h!]
\centering
\begin{subfigure}{0.5\textwidth}
    \includegraphics[width=\textwidth]{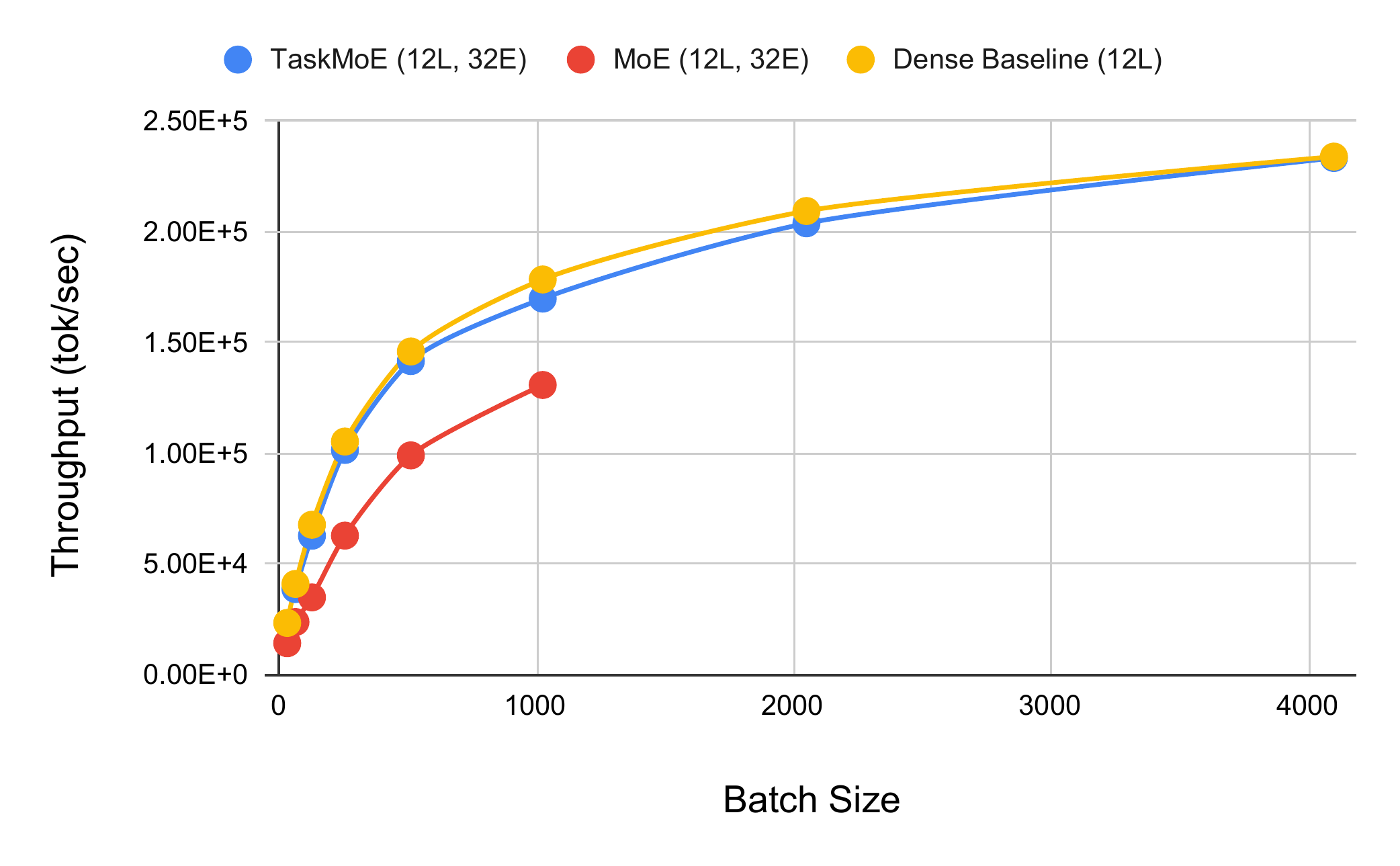}
\end{subfigure}
\caption{\textbf{Inference cost analysis:} We measure the throughput of our Task-MoE model, baseline Transformer-Base model and baseline Token-MoE model across batch sizes and see that the peak throughput of Task-MoE (and Transformer-Base) is 1.87 times higher than that of Token-MoE.}
\label{fig:inf-compare}
\end{figure}

We further compare Task-level MoEs with Token-level MoEs in terms of throughput across different batch sizes in Figure \ref{fig:inf-compare}.  We measure this by decoding the WMT14 English-German test set with our TaskMoE model and with the baseline TokenMoE model on 32 Cloud TPU V3 cores. We find that our Task-MoE model has 1.87 times higher peak throughput while using 3.75 times less decoder parameters (142M vs 533M). Moreover, our Task-MoE model has minimal communication overhead compared to decoding with Token-MoE ($0.0\%$ versus $26.9\%$ of step time). 

We note that the inference time of the token-based MoE model is dominated by the decoder, with the decoders taking 200x the time per step than the encoders at peak throughput. Therefore, the inference cost of task-level routing on decoder only is roughly equivalent to that on both the encoder and decoder.

\subsection{Comparison of Extracting Task MoE Models to Distillation}\label{sec:wmt-distill}

While in Section \ref{sec:wmt-throughput} we compared the throughput of task-level MoE and token-level MoE models, it is common  practice for large models to be distilled to smaller student models suitable for deployment.

We distill our token-level MoE baseline to Transformer-Base student models with the same architecture as the multlingual dense baseline discussed in \ref{sec:wmt-setup}. As done in \cite{DBLP:journals/corr/abs-2101-03961}, we initialize the student model with non-expert weights of the teacher model. We distill the model with the source sides of the WMT parallel data used while training the original teacher model. We do this for several language pairs across different language families and resource sizes - EnFr, FrEn, DeEn, EnDe, EnRo, RoEn, EnHi and HiEn. Additional training details are provided in the Appendix \ref{sec:wmt-model-details}. 

In Table \ref{tab:distill-compare}, we compare the BLEU scores of our best performing Task MoE models to distillation of our Token MoE baseline into models with similar inference cost (shown in Figure \ref{fig:inf-compare}).  We see that distillation consistently underperforms our best-performing Task MoE model - distillation from Token MoE achieves an average BLEU score of 26.9, while our best-performing Task MoE model has an average BLEU score of 29.0 (+2.1 BLEU) for these language pairs. We note that while distilling our sparse MoE model, only 32.25\% of gains over dense multilingual baselines are preserved. This is in line with the distillation results discussed in \cite{DBLP:journals/corr/abs-2101-03961}.

\section{Scaling up to 200 Language Pairs}\label{sec:m4-exp}

We now scale our results up to a larger internal dataset with over 200 language pairs, while also scaling the number of parameters to beyond 10 billion weights. In addition, we look more closely at the gating decisions made by these sparse models and discuss their implications. 

\subsection{Experimental Setup}\label{sec:m4-setup}
\paragraph{Data:} We use an in-house training corpus generated by crawling and extracting parallel sentences from the web \citep{uszkoreit2010large}. This dataset has 204 direct language pairs (102 languages to and from English), with a total of 25 billion sentence pairs. This dataset covers a diverse range of domains and languages, and is quite noisy. There is also a heavy imbalance when it comes to the number of examples available per language pair, ranging between $10^4$ and $10^9$ sentence pairs.  In order to record gating decisions while controlling for semantics, we created a multi-way aligned evaluation set containing nearly 3k sentence pairs for all languages.\footnote{Each sentence in our evaluation set is semantically identical across all other languages.}

\begin{figure}[t!]
\centering
\begin{subfigure}{0.5\textwidth}
    \centering
    \includegraphics[width=\textwidth]{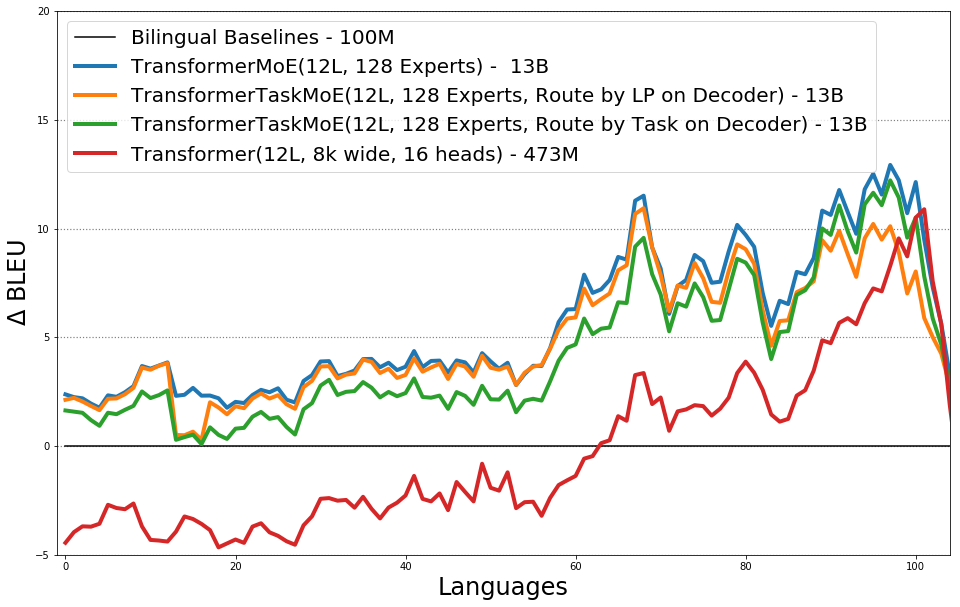}
    \caption{Performance of different routing strategies on \textit{Xx-En} language pairs.}
    \label{fig:xen_m4}
\end{subfigure}

\begin{subfigure}{0.5\textwidth}
    \centering
    \includegraphics[width=\textwidth]{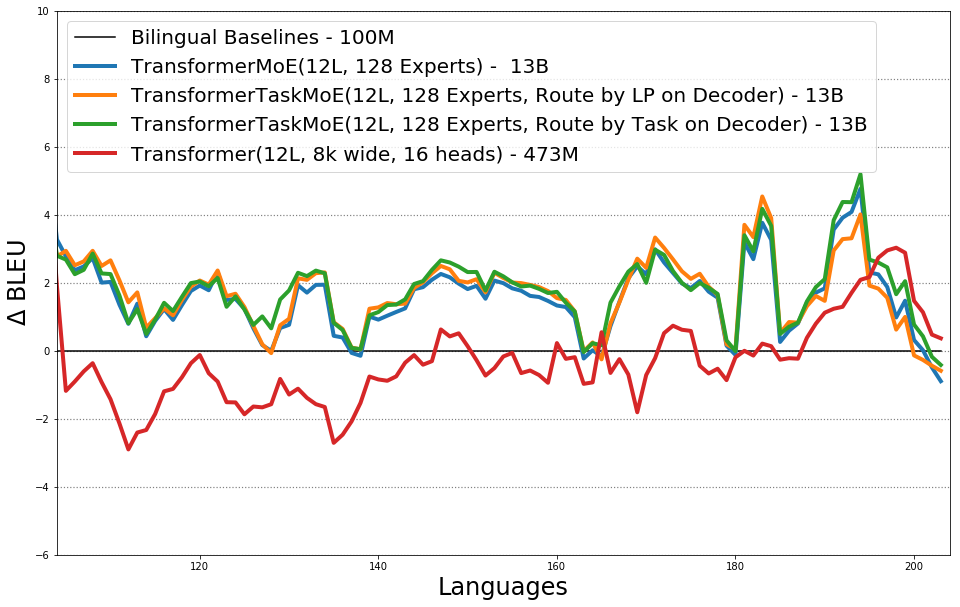}
    \caption{Performance of different routing strategies on \textit{En-Xx} language pairs.}
    \label{fig:enx_m4}
\end{subfigure}
\caption{{\bf Comparing the performance of different routing strategies for Mixture-of-Experts (MoE) models on a massively multilingual dataset} -- We compare routing experts by tokens, and tasks (using either language pairs or target languages). Given that routing by token on the encoder and routing by task on the decoder performed the best on WMT (Table \ref{tab:32exp-compare}), we use those settings for the scaled up 128 expert models we compare. We split the comparison of results into (a) \textit{Xx-En} language pairs and (b) \textit{En-Xx} language pairs. The languages on the x-axis are sorted left-to-right in descending order of resource size. Best seen in color. Note that the token-level MoE has $6.5$B parameters in the decoders while our task-level MoE has only $200$M.}
\label{fig:m4-res}
\end{figure}

\paragraph{Model:} We use the 473M Transformer Big \citep{vaswani2017attention} architecture (or modified versions of it in the case of sparse models) as described by \cite{chen-EtAl:2018:Long1} for this set of experiments. Similar to Section \ref{sec:wmt-setup}, we (1) share all parameters across language pairs including softmax layer and input/output word embeddings, (2) pre-pend a \verb <2xx>  token to the source sentence to indicate the target language and (3) use a Sentence Piece Model \cite{kudo2018sentencepiece} with 64k tokens vocabulary shared on both the encoder and decoder side.We followed the training and architecture as shown in \citet{lepikhin2020gshard}.\footnote{As opposed to displaying BLEU scores for each language pair, we place the baselines on the $x$-axis at zero and report the $\Delta$BLEU trendline of each model we consider. In order to set these bilingual baselines, we train Neural Machine Translation models for each language pair (e.g. a single model for German-to-English), tuned depending on the available training data for that given language We tuned batch-size and different values of regularization methods (e.g. dropout) in a Transformer-Big or Transformer-Base layout, for high or low-resourced languages respectively.} 


\subsection{Results}\label{sec:m4-res2}

We compare Task-level MoEs and Token-level MoEs to their bilingual and multilingual baselines in Figure 2.
We train 128 expert MoE models with routing in these settings: (1) Routing by token on both the encoder and decoder, (2) Routing by token on the encoder and by target language on the decoder and (3) Routing by token on the encoder and by language pair on the decoder.

We find that these scaled up sparse models perform better than their dense baselines, with hybrid task-level routing performing slightly better on \textit{En-Xx} language pairs and pure token-level routing performing slightly better on \textit{Xx-En} language pairs. We hypothesize that for the \textit{Xx-En} tasks, not explicitly dividing expert parameters by tasks on the decoder results in better transfer, thus explaining the better performance of token-level routing. This suggests that a hybrid strategy that partially restricts access to experts based on task-boundaries, while still permitting routing by tokens, might provide the right balance between efficiency and quality.

We also note that while both forms of routing have 13B parameters (6.5B on decoder) at train time, token level routing only on the decoder uses only 200M parameters at inference time, in addition to the practical considerations discussed in Section \ref{sec:routing}. We provide aggregate BLEU scores in Appendix \ref{sec:m4-bleu} and parameter count breakdowns in Appendix \ref{sec:param-count-m4}. In addition, we take a closer look at routing decisions made for different languages by the model in Section \ref{sec:moe-analysis}.



\subsection{Comparison of Throughput on Massive Models}\label{sec:compare-throughput}

\begin{figure}[h!]
\centering
\begin{subfigure}{0.45\textwidth}
    \includegraphics[width=\textwidth]{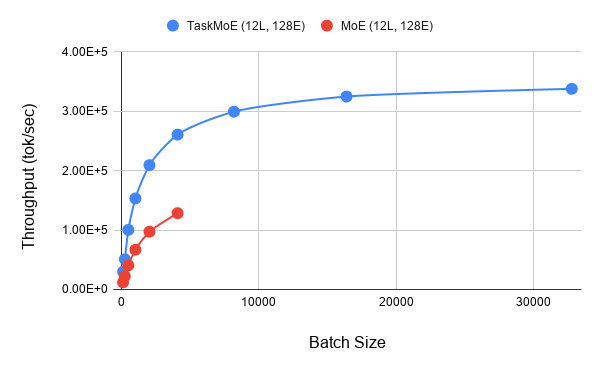}
\end{subfigure}
\caption{\textbf{Inference cost analysis:} We measure the throughput of our Task-MoE model and baseline Token-MoE model across batch sizes and see that the peak throughput of Task-MoE is 2.6 times higher. }
\label{fig:moe-throughput}
\end{figure}

Similar to Section \ref{sec:wmt-throughput}, we compare Task-level MoEs with Token-level MoEs in terms of throughput across different batch sizes in Figure \ref{fig:moe-throughput}.  We decode the WMT14 English-German test set with our TaskMoE model and with the baseline TokenMoE model on 128 Cloud TPU V3 cores. We find that our Task-MoE model has 2.6 times higher peak throughput while using 32.34 times less decoder parameters (201M vs 6.5B). Moreover, our Task-MoE model has minimal communication overhead compared to decoding with Token-MoE ($0.2\%$ versus $36\%$ of step time).

\subsection{A Closer Look at the Routing Decisions}\label{sec:moe-analysis}

\begin{figure*}[h!]

\begin{subfigure}{\textwidth}
    \centering
    \includegraphics[width=0.75\textwidth]{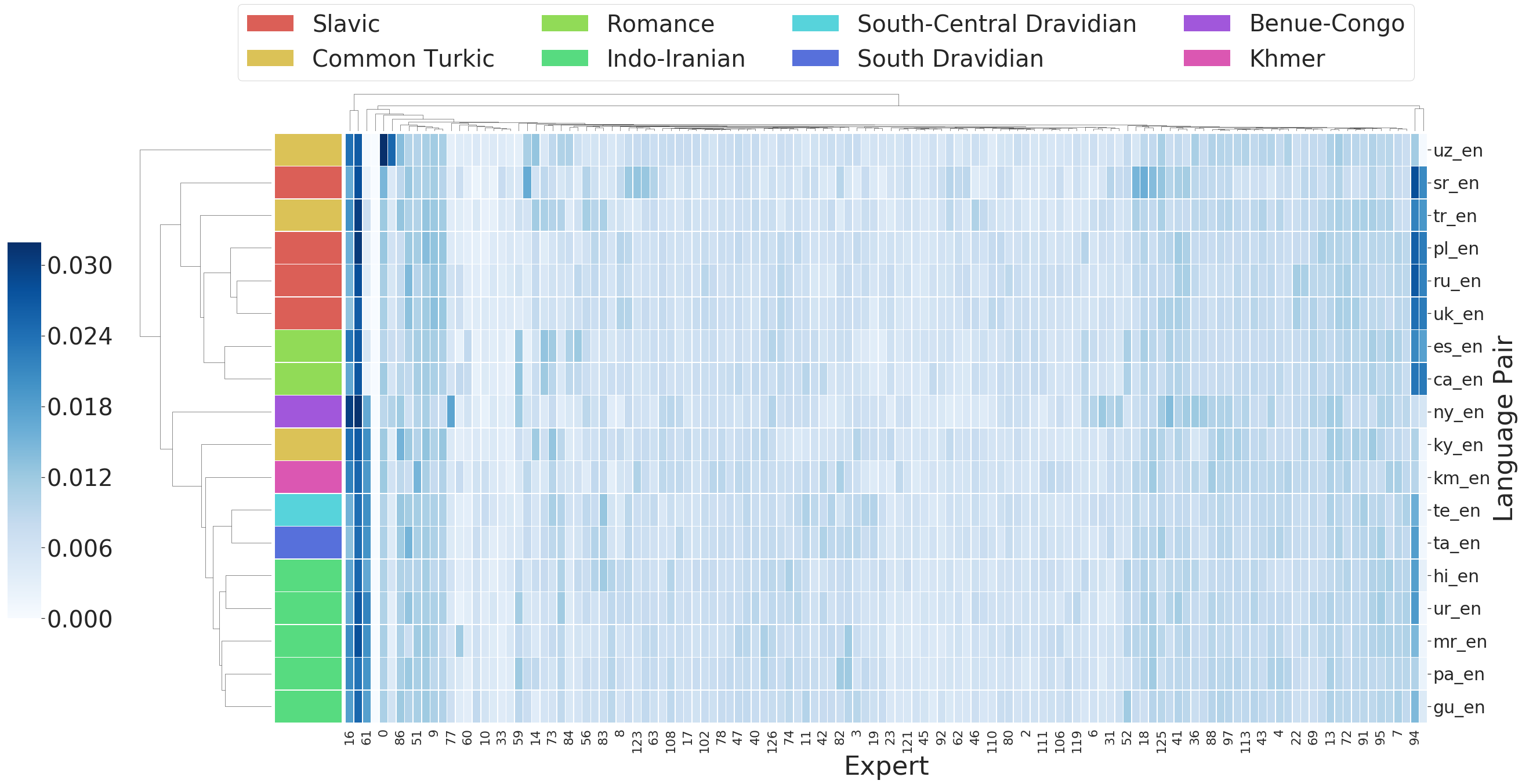}
    \caption{Gating decisions of the last layer of the encoder for Xx-En language pairs.}
    \label{fig:enc-layer6}
\end{subfigure}

\begin{subfigure}{\textwidth}
    \centering
    \includegraphics[width=0.75\textwidth]{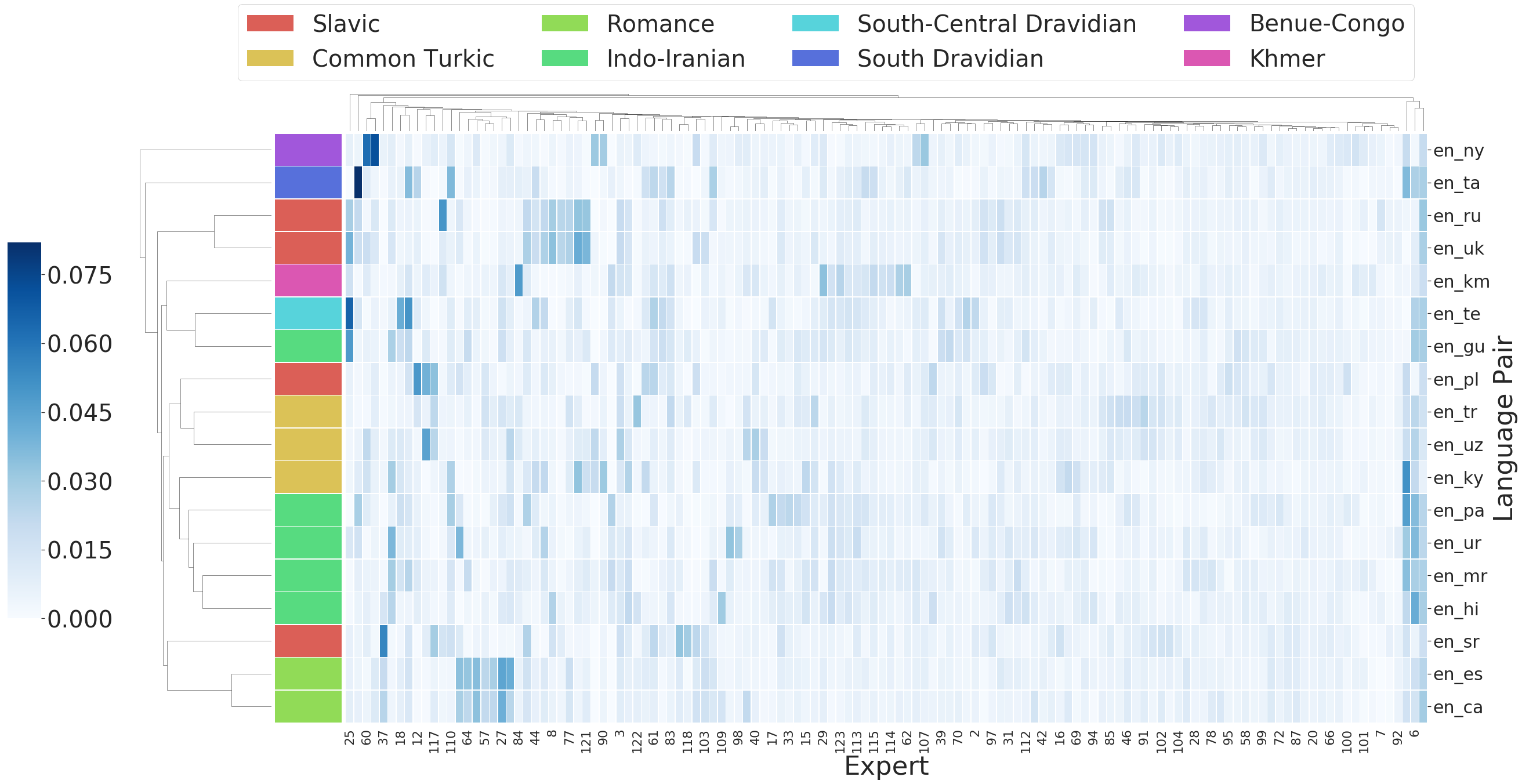}
    \caption{Gating decisions of the last layer of the decoder for En-Xx language pairs.}
    \label{fig:dec-layer6-enx}
\end{subfigure}
\caption{We record the gating decisions of our MoE model trained on internal data on a multiway parallel dataset. The darker a cell, corresponding to, say en-sr and the 37th expert, the more the expert is used.  In (a) the encoder, tokens from all tasks (\textit{Xx-En}) seem to prefer the same set of few experts slightly over the others; while in (b) the decoder each task (\textit{En-Xx}) seems to slightly prefer a few experts over the other. Moreover, the set of experts appears to be similar for related languages. For example, English-Spanish and English-Catalan (two Romance Languages) have similar expert distributions and so do English-Russian and English-Ukranian (two Slavic Languages).}
\label{fig:moe-gating-main}
\vspace{-0.1in}
\end{figure*}

Now, we analyze the routing decisions made in token-level MoE models to further motivate our investigation. We take a token-level MoE model  trained on the massively multilingual dataset and decode these models on the multiway test-sets, while logging routing decisions for every token. We plot the top expert distributions of several tasks with different scripts and language families in Figure \ref{fig:moe-gating-main}. For clarity, and because these two groups of languages behave differently in a multilingual setting, we split the gating decisions into those for \textit{Xx-En} and \textit{En-Xx} language pairs. In the encoder (Figure 5a), tokens from all tasks (\textit{Xx-En}) seem to prefer the same set of few experts slightly over the others. On the other hand, in the decoder (Figure 5b) each task seems to have a slight preference for a few experts over the others. Moreover, the set of experts appears to be similar for related languages. For example, English-Spanish and English-Catalan (two Romance Languages) have similar expert distributions and so do English-Russian and English-Ukranian (two Slavic Languages). In the Appendix \ref{sec:moe-analysis-sup}, we provide expert distribution plots for other layers of this model. In addition, we provide expert distributions of the MoE model that routes tokens by target language discussed in Section \ref{fig:m4-res}.

Our analysis suggest that, when using token-level routing, task-level decisions emerge naturally in the decoder, providing additional motivation for our proposed routing strategies.

\section{Related Work}

\paragraph{Conditional Computation:} Conditional computation \cite{bengio2015conditional}, or routing examples through the neural network by activating only a sub-network of the network depending on the input has seen success in large scale natural language processing (NLP) (\cite{shazeer2017outrageously,lepikhin2020gshard,bapna2019simple}) and computer vision (\cite{yang2019condconv}) tasks. A variety of strategies can be used to route examples such as learning a function on the input \cite{shazeer2017outrageously,lepikhin2020gshard}, computational budget \cite{bapna2019simple,elbayad2019depth} or simplifying the expert allocation and training regimen \cite{DBLP:journals/corr/abs-2103-16716, DBLP:journals/corr/abs-2101-03961}. 

\paragraph{Multi-task Learning} Multi-task Learning improves model performance across all tasks trained on due to regularization and positive transfer between related tasks~\cite{caruana1997multitask}. Here, sub-networks are be activated depending on the task to which the input belongs - some of these parameters may be shared. This approach has seen success in a variety of domains such as classification, recommender systems and NLP (\cite{ma2019snr,ma2018modeling,clark2019bam,collobert2008unified,ruder2019latent,tan2019multilingual}). Like our work, some of these models have been designed with inference benefits in mind (\cite{ma2019snr}). In this work we focus on multi-task learning in the case of Multilingual NMT.

\paragraph{Multi-task learning for Multilingual NMT Models:} Multi-task learning in multilingual models has been well-studied: while complete parameter sharing is simple and works well (\cite{johnson2017google}), an optimal strategy for sharing parameters and possibly having languages-specific parameters would maximize transfer while minimizing interference \cite{hokamp2019evaluating}. Strategies involve allocating language specific hidden states,  attention modules, decoders or additional specialized layers (\cite{hokamp2019evaluating,wang2018three,gu2018universal,bapna2019simple}). In addition some strategies involve grouping parameters by language group \cite{fan2020beyond,tan2019multilingual}. Compared to these works, our approach to parameter sharing is designed to scale models without impacting inference efficiency (as opposed to simply adding language-specific capacity) while still enjoying the benefits of scaling.
Most similar to our work in terms of the inference utility is proposed by \cite{li2020deep} where discrete latent variables used to learn language specific layer combinations, whereas in our study we focus on improving inference efficiency of mixture of expert models at scale. 

\section{Conclusions}

In this work we discussed more inference friendly algorithms for routing examples in multilingual Sparse Mixture-of-Experts models by making use of task boundaries. We empirically demonstrated that this new algorithm performs as well as, or better than, conventional token-based routing algorithms on two different datasets: a multilingual WMT setup covering 30 language pairs and a large internal dataset covering 200 language pairs, in terms of machine translation quality evaluated with BLEU.
By carefully comparing inference throughput across different routing approaches and distilled models, we demonstrated the superiority of task-based routing algorithms over either serving a  token-based MoE model as-is (in terms of peak throughput) and over distilling  a large MoE model into a smaller dense model (in terms of BLEU). 

We conclude by highlighting that algorithms that are more inference friendly while retaining the quality gains of MoE models are a promising direction for future exploration, motivating research on \textit{inference efficiency} for large models. Although we studied some hybrid routing strategies where encoder and decoder networks utilize different routing schemes, we believe that future research on more granular routing hybrids or hierarchical variants will deliver more gains and advance our understanding of large scale, sparsely gated, massively multi-task networks.

\section{Acknowledgements}
We would like to thank Wolfgang Macherey, Yuanzhong Xu and Macduff Richard Hughes for their helpful feedback on the draft. We would also like to thank the Google Translate and Google Brain teams for their useful input and discussions, and the entire GShard development team for their foundational contributions to this project. In addition, we thank the anonymous reviewers for their insightful comments.



\bibliography{emnlp2021}
\bibliographystyle{acl_natbib}

\pagebreak

\appendix

\section{Appendix}\label{sec:app}

\subsection{WMT Model and Training Details}
\label{sec:wmt-model-details}


For our experiments, we use the Transformer Base model in \citep{chen-EtAl:2018:Long1}, The sole difference is that we use a 64k vocabulary: our model therefore contains 142M parameters. For multilingual models, we share all parameters across language pairs including softmax layer in input/output word embeddings. 

We use a 64k token vocabulary formed using a Sentence Piece Model \citep{kudo2018sentencepiece}. The vocabulary is shared on both the encoder and decoder side. To learn a joint SPM model given our imbalanced dataset, we followed the temperature based sampling strategy with a temperature of $T=5$.

Finally, our models are optimized using the Adafactor optimizer \citep{shazeer2018adafactor} with momentum factorization and a per-parameter norm clipping threshold of 1.0. We followed a learning rate of of 3.0,  with 40K warm-up steps for the schedule, which is decayed with the inverse square root of the number of training steps after warm-up. BLEU scores presented in this paper are calculated using SacreBLEU \cite{post-2018-call} on the WMT test sets. \footnote{ \texttt{ BLEU+case.mixed+lang.<sl>-<tl>+
numrefs.1+smooth.exp+tok.<tok>+version
.1.3.0} ,   
where $sl$ is the source language, $tl$ is the target language and $tok=zh$ if $tl=zh$ and $intl$ otherwise.}

For distillation, training and model details are identical apart from a reduced learning rate of 0.2.

\subsection{WMT Dataset Details}\label{sec:wmt-data}

In Table \ref{tab:dataset-bi} we provide the training set details for the WMT \footnote{\url{http://www.statmt.org/wmt20/}} setup we use  \citep{siddhant2020leveraging}. We provide the data sizes and WMT years of the Train, Dev and Test sets we use.

\begin{table*}[h]
\centering
\begin{tabular}{>{\centering\arraybackslash}m{2cm}>{\centering\arraybackslash}m{1.7cm}>{\centering\arraybackslash}m{1.7cm}>{\centering\arraybackslash}m{1.7cm}>{\centering\arraybackslash}m{1.7cm}>{\centering\arraybackslash}m{1.2cm}>{\centering\arraybackslash}m{1.2cm}}
\toprule
 \multirow{2}{1.5cm}{\centering Language Pair} & \multicolumn{3}{c}{Data Sources} & \multicolumn{3}{c}{$\#$ Samples}\\
 \cmidrule{2-7}
  & Train & Dev & Test & Train & Dev & Test\\
\midrule
cs$\rightarrow$en                                                            & WMT'19    & WMT'17    & WMT'18   & 64336053     & 3005    & 2983    \\
fr$\rightarrow$en                                                            & WMT'15    & WMT'13    & WMT'14   & 40449146     & 3000    & 3003    \\
ru$\rightarrow$en                                                            & WMT'19    & WMT'18    & WMT'19   & 38492126     & 3000    & 2000    \\
zh$\rightarrow$en                                                            & WMT'19    & WMT'18    & WMT'19   & 25986436     & 3981    & 2000    \\
es$\rightarrow$en                                                            & WMT'13    & WMT'13    & WMT'13   & 15182374     & 3004    & 3000    \\
fi$\rightarrow$en                                                            & WMT'19    & WMT'18    & WMT'19   & 6587448      & 3000    & 1996    \\
de$\rightarrow$en                                                            & WMT'14    & WMT'13    & WMT'14   & 4508785      & 3000    & 3003    \\
et$\rightarrow$en                                                            & WMT'18    & WMT'18    & WMT'18   & 2175873      & 2000    & 2000    \\
lv$\rightarrow$en                                                            & WMT'17    & WMT'17    & WMT'17   & 637599       & 2003    & 2001    \\
lt$\rightarrow$en                                                            & WMT'19    & WMT'19    & WMT'19   & 635146       & 2000    & 1000    \\
ro$\rightarrow$en                                                            & WMT'16    & WMT'16    & WMT'16   & 610320       & 1999    & 1999    \\
hi$\rightarrow$en                                                            & WMT'14    & WMT'14    & WMT'14   & 313748       & 520     & 2507    \\
kk$\rightarrow$en                                                            & WMT'19    & WMT'19    & WMT'19   & 222424       & 2066    & 1000    \\
tr$\rightarrow$en                                                            & WMT'18    & WMT'17    & WMT'18   & 205756       & 3007    & 3000    \\
gu$\rightarrow$en                                                            & WMT'19    & WMT'19    & WMT'19   & 155798       & 1998    & 1016    \\
\midrule
en$\rightarrow$cs                                                            & WMT'19    & WMT'17    & WMT'18   & 64336053     & 3005    & 2983    \\
en$\rightarrow$fr                                                            & WMT'15    & WMT'13    & WMT'14   & 40449146     & 3000    & 3003    \\
en$\rightarrow$ru                                                            & WMT'19    & WMT'18    & WMT'19   & 38492126     & 3000    & 2000    \\
en$\rightarrow$zh                                                            & WMT'19    & WMT'18    & WMT'19   & 25986436     & 3981    & 2000    \\
en$\rightarrow$es                                                            & WMT'13    & WMT'13    & WMT'13   & 15182374     & 3004    & 3000    \\
en$\rightarrow$fi                                                            & WMT'19    & WMT'18    & WMT'19   & 6587448      & 3000    & 1996    \\
en$\rightarrow$de                                                            & WMT'14    & WMT'13    & WMT'14   & 4508785      & 3000    & 3003    \\
en$\rightarrow$et                                                            & WMT'18    & WMT'18    & WMT'18   & 2175873      & 2000    & 2000    \\
en$\rightarrow$lv                                                            & WMT'17    & WMT'17    & WMT'17   & 637599       & 2003    & 2001    \\
en$\rightarrow$lt                                                            & WMT'19    & WMT'19    & WMT'19   & 635146       & 2000    & 1000    \\
en$\rightarrow$ro                                                            & WMT'16    & WMT'16    & WMT'16   & 610320       & 1999    & 1999    \\
en$\rightarrow$hi                                                            & WMT'14    & WMT'14    & WMT'14   & 313748       & 520     & 2507    \\
en$\rightarrow$kk                                                            & WMT'19    & WMT'19    & WMT'19   & 222424       & 2066    & 1000    \\
en$\rightarrow$tr                                                            & WMT'18    & WMT'17    & WMT'18   & 205756       & 3007    & 3000    \\
en$\rightarrow$gu                                                            & WMT'19    & WMT'19    & WMT'19   & 155798      & 1998    & 1016    \\
\midrule
fr$\rightarrow$de                                                            & WMT'19    & WMT'13    & WMT'13   & 9824476      & 1512    & 1701    \\
de$\rightarrow$fr                                                            & WMT'19    & WMT'13    & WMT'13   & 9824476      & 1512    & 1701    \\
\bottomrule
\end{tabular}
\caption{Data sources and number of samples for the parallel data in our corpus. Please note that we don't use parallel data in Fr-De for any of the experiments in the paper.} 

\label{tab:dataset-bi}
\end{table*}

\subsection{Individual WMT BLEU Scores}\label{sec:wmt-full-res}

\textbf{Bilingual baselines:} We first train Transformer Base and Big models on each language pair. The results are in Table \ref{tab:bilingual}.
\begin{table*}[t]
\centering
\resizebox{\linewidth}{!}{
\begin{tabular}{c|ccccccccccccccc}
\toprule
xx                  & cs   & fr   & ru   & zh   & es   & fi   & de   & et   & lv   & lt   & ro   & hi  & kk  & tr   & gu  \\ \midrule
Any-to-English (xx$\rightarrow$en) & 31.3 & 37.2 & 36.0 & 21.7 & 32.7 & 27.3 & 31.7 & 23.1 & 15.0 & 21.3 & 30.1 & 8.5 & 11.5 & 15.9 & 1.0 \\
English-to-Any (en$\rightarrow$xx) & 23.8 & 41.3 & 26.4 & 31.3 & 31.1 & 18.1 & 29.9 & 18.2 & 14.2 & 11.5 & 23.4 & 4.5 & 1.9  & 13.6 & 0.6 \\ \bottomrule
\end{tabular}} 
\caption{Bilingual baselines. xx refers to language in the column header. \citep{siddhant2020leveraging}} \label{tab:bilingual}
\end{table*}

In Tables 5 and 6
we provide individual BLEU scores of the models discussed in Table \ref{tab:32exp-compare}.


\begin{sidewaystable}[]
\resizebox{\linewidth}{!}{
\begin{tabular}{c|c|c|c|cc|cc|cc|cc|cc|cc|cc|cc|}
\toprule
\multirow{2}{*}{System}                           & \multicolumn{2}{c|}{Routing Granularity} & \multicolumn{17}{c}{BLEU}  \\                                                                          \cline{2-20}                                                                                                                     
                                                  &                        &                & AVG   & xx2en & en2xx & HRL   & LRL    & cs\_en & en\_cs & fr\_en & en\_fr & ru\_en & en\_ru & zh\_en & en\_zh & es\_en & en\_es & de\_fr & fr\_de \\
                                                  
                                                  \hline

Multilingual Transformer-Base   &    -   &     -      & 20.03 & 23.69 & 17.5  & 23.25 & 15.88  & 27.2   & 18.1   & 34.1   & 36.1   & 31.7   & 21.1   & 18.9   & 17.2   & 31.3   & 29.2   & 17.4   & 5.5  \\
Multilingual Transformer-Big   &    -   &     -      & 23.84   & 26.10  & 22.03   & 27.69   & 18.89 & 31.03 & 23.24 & 37.75  & 40.43 & 35.2  & 25.09   & 20.02   & 25.99   & 33.45   & 32.27   & 20.07   & 20.98     \\
\hline
Sentence-level MoE -- 32 expert      &   Sentence    & Sentence         & 19.88 & 24.05 & 16.83 & 22.56 & 14.14 & 27.6   & 18.7   & 34.4   & 36.5   & 32.7   & 15.1   & 20.4   & 7.2    & 31.3   & 30.1   & 13.6   & 9.1    \\
\hline
Token-level MoE -- 32 experts     &   Token    &     Token         & 22.58 & 24.91 & 20.35 & 27.49 & 16.28  & 29.8   & 21.8   & 36.4   & 40.1   & 34.6   & 25.7   & 19.9   & 23.7   & 33.9   & 32.8   & 23.9 & 19.9  \\
\hline
\multirow{6}{*}{Task-level MoE -- 32 experts}       &    Language Pair  &     Language Pair        & 22.04 & 25.43 & 19.5  & 25.57 & 17.5   & 26.8   & 21.7   & 35.4   & 39.2   & 33     & 21     & 22.1   & 17.9   & 32.4   & 32.1   & 12.2   & 19.1 \\
 &         Target    &    Target         & 22.88 & 25.63 & 20.19 & 27.21 & 17.3   & 29.1   & 21.7   & 36.1   & 40.2   & 33.8   & 24.7   & 21.9   & 24.8   & 32.6   & 33.1   & 25.8   & 18.8 \\
          &           Language Pair   &      Token        & 22.45 & 25.58 & 20.34 & 26.85 & 16.79  & 30.3   & 21.5   & 36.7   & 40.3   & 34.8   & 25.1   & 21     & 25.9   & 33.6   & 32.4   & 12.9   & 16.6 \\
          &              Target   &      Token          & 22.33 & 24.47 & 20.44 & 26.82 & 16.55  & 29.4   & 22     & 35.3   & 39.7   & 33.8   & 25.2   & 21     & 26.2   & 32.4   & 32.7   & 22.2   & 18.6 \\
  &             Token  &  Language Pair        & 23.03 & 26.16 & 20.28 & 27.23 & 17.62  & 30.1   & 23.2   & 37.5   & 39.5   & 35.5   & 21.9   & 21.7   & 15.7   & 34.5   & 33.5   & 20.1   & 20.1  \\
     &         Token     &    Target     & 23.62 & 25.95 & 21.09 & 28.48 & 17.37  & 30.5   & 22.5   & 37.1   & 39.9   & 35.4   & 25.6   & 21.4   & 27     & 34.3   & 33.5   & 27.7   & 22.4 \\

\bottomrule
\end{tabular}
}

\caption{Part 1 of the table with individual BLEU scores for Table\ref{tab:32exp-compare}}
\end{sidewaystable}\label{tab:32exp-full}

\begin{sidewaystable}[]
\resizebox{\linewidth}{!}{
\begin{tabular}{c|c|c|cc|cc|cc|cc|cc|cc|cc|cc|cc|cc}
\toprule
\multirow{2}{*}{System}                           & \multicolumn{2}{c|}{Routing Granularity} & \multicolumn{18}{c}{BLEU}  \\                                                                          \cline{2-21}                                                                                                                     
                                                  &                        &                & fi\_en & en\_fi & de\_en & en\_de & et\_en & en\_et & lv\_en & en\_lv & lt\_en & en\_lt & ro\_en & en\_ro & hi\_en & en\_hi & kk\_en & en\_kk & tr\_en & en\_tr & gu\_en & en\_gu \\
                                                  
                                                  \hline

Multilingual Transformer-Base   &    -   &     -     & 23.9    & 17     & 28.6   & 22     & 23.1   & 16.1   & 17.2   & 14.9   & 24.6   & 11.4   & 33.4   & 23.9   & 19.2   & 10.4   & 13.5   & 2.5    & 20.9   & 17.5   & 7.8    & 5.1    \\

Multilingual Transformer-Big   &    -   &     -     & 27.89    & 20.83    & 30.72   & 27.37     & 28.49   & 17.59   & 20.32   & 17.76   & 26.1   & 26.1   & 35.84   & 26.83   & 20.87   & 14.61   & 10.4   & 5.23    & 22.69   & 19.44   & 10.68    & 7.67    \\
\hline
Sentence-level MoE -- 32 expert      &   Sentence    & Sentence         23.5   & 17.2   & 29.4   & 21.8   & 22     & 15.4   & 17.9   & 14.7   & 24.6   & 11.6   & 33.6   & 24.8   & 20.5   & 12.2   & 14     & 2.9    & 21.4   & 17.9   & 7.4    & 6.3    \\
\hline
Token-level MoE -- 32 experts     &   Token    &     Token     & 27.3   & 20.2   & 31.2   & 26.7   & 27     & 19.9   & 18.7   & 17     & 23.7   & 13.9   & 33.7   & 26.5   & 19.8   & 11.5   & 8.5    & 2.4    & 20.3   & 18     & 8.8    & 5.1    \\
\hline
\multirow{6}{*}{Task-level MoE -- 32 experts}       &    Language Pair  &     Language Pair       & 25.2   & 20.1   & 31.3   & 26.9   & 24.7   & 19.2   & 18.4   & 16.3   & 25.1   & 13.6   & 34.8   & 25.7   & 22.5   & 13.1   & 15     & 2.4    & 23.4   & 18.2   & 11.4   & 5.1    \\
 &         Target    &    Target    & 25.6   & 19.5   & 30.7   & 26.8   & 24.8   & 19.8   & 18.4   & 15.7   & 25.9   & 13.6   & 34.9   & 25.8   & 21.7   & 12.3   & 15.5   & 2.4    & 22.5   & 17.7   & 11     & 4.8    \\
          &           Language Pair   &      Token   & 26.7   & 20     & 32.2   & 26.9   & 26.8   & 19.6   & 18.9   & 16.3   & 25.1   & 13.3   & 34.2   & 25.8   & 21.1   & 12.6   & 12.6   & 2.3    & 21.7   & 18.4   & 8      & 4.7    \\
          &              Target   &      Token      & 23.7   & 19.8   & 30.7   & 26.1   & 24.1   & 19.9   & 18     & 16.5   & 24.4   & 13.6   & 33.1   & 26.1   & 20     & 12.7   & 12.7   & 2.9    & 21.1   & 18.2   & 7.4    & 5      \\
  &             Token  &  Language Pair   & 27.8   & 21.1   & 32.3   & 27     & 27.6   & 21     & 19.8   & 17.2   & 26     & 14.6   & 36.4   & 26.8   & 20.4   & 14.2   & 12.3   & 3.3    & 21.5   & 19.4   & 9      & 5.8    \\
     &         Token     &    Target   & 27.9   & 20.5   & 32     & 27.1   & 27.3   & 20.5   & 19.4   & 17.6   & 25.9   & 14.4   & 36.2   & 26.6   & 20.1   & 13.3   & 11.6   & 3      & 21.2   & 19.2   & 9      & 5.7   \\

\bottomrule
\end{tabular}
}
\caption{Part 2 of the table with individual BLEU scores for Table\ref{tab:32exp-compare}}
\end{sidewaystable}\label{tab:32exp-full-2}

\subsection{Detailed Breakdown of Parameter Counts on WMT}\label{sec:param-count-wmt}

Table 7 
describes the parameter counts of different parts of the Transformers compared in Table \ref{tab:32exp-compare}.

\begin{sidewaystable}[]
\resizebox{\linewidth}{!}{
\begin{tabular}{c|c|c|c|c|c|c|c|c|c|c}
\toprule
\multirow{2}{*}{{\bf System}} & \multicolumn{2}{c|}{{\bf Routing Granularity}} & \multicolumn{5}{c|}{{\bf No. of Parameters}}  &\multicolumn{3}{c}{{\bf Effective n(params) at inference time}}  \\
\cline{2-11}
& Encoder & Decoder & Vocabulary & Encoder  & Decoder & Softmax & Total  &  Encoder  & Decoder  & Total  \\
\hline

Multilingual Transformer-Base   &    -   &     -   & 33M  &  19M  &  25M & 65M & 142M & 19M & 25M & 142M \\
\hline
Token-level MoE -- 32 experts     &   Token    &     Token   &\multirow{8}{*}{33M} & \multirow{8}{*}{214M} & \multirow{8}{*}{221M}& \multirow{8}{*}{65M} & \multirow{8}{*}{533M} & 214M & 221M & 533M \\

Sentence-level MoE -- 32 expert      &   Sentence    & Sentence  & & & & & & 214M & 221M & 533M\\

\multirow{6}{*}{Task-level MoE -- 32 experts} &    Language Pair  &     Language Pair     & & & & & & 25M & 32M & 155M   \\
 &   Target    &    Target   &  & & & & & 25M & 32M & 155M   \\
 &    Language Pair   &      Token  & & & & & & 214M & 25M & 338M   \\
 &    Target   &      Token  & & & &	&  &	 214M  &	25M &	338M \\
 &  Token  &  Language Pair & & & & & & 19M & 221M & 338M \\
 &  Token     &    Target &  & & & & & 19M & 221M  & 338M \\
\bottomrule
\end{tabular}
}
\caption{We break down the parameter counts of the models we compare in Section \ref{sec:routing-results} by components.}
\end{sidewaystable}\label{tab:param-count-wmt}





\subsection{Detailed Breakdown of Parameter Counts}\label{sec:param-count-m4}

In Table 8
we describe the parameter counts of different parts of the Transformers discussed in Section \ref{sec:m4-exp}.

\begin{sidewaystable}[]

\resizebox{\linewidth}{!}{
\begin{tabular}{c|c|c|c|c|c|c|c|c|c|c}
\toprule
\multirow{2}{*}{{\bf System}} & \multicolumn{2}{c|}{{\bf Routing Granularity}} & \multicolumn{4}{c|}{{\bf No. of Parameters}}  &\multicolumn{3}{c}{{\bf Effective n(params) at inference time}}  \\
\cline{2-11}
& Encoder & Decoder & Vocabulary & Encoder  & Decoder & Softmax & Total  &  Encoder  & Decoder  & Total  \\
\hline
Multilingual Transformer-Big   &    -   &     -   & \multirow{4}{*}{{ 65M}}   &  126M  &  151M & \multirow{4}{*}{{ 131M}}   & 473M  &  126M  &  151M & 473M \\
Token-level MoE -- 128 experts     &   Token    &     Token   &  &  6.5B  & 6.5B  &   & 13B & 6.5B  & 6.5B   & 13.3B \\

Task-level MoE -- 128 experts      &   Token    & Language & & 6.5B   & 6.5B &    & 13B & 6.5B  & 201M   & 6.9B  \\
Task-level MoE -- 128 experts      &   Token    & Target &  & 6.5B   & 6.5B  &  & 13B & 6.5B  & 201M   & 6.9B   \\

\bottomrule
\end{tabular}
}
\caption{We break down the parameter counts of the models we compare in Section \ref{sec:m4-res2} by components.}
\end{sidewaystable}\label{tab:param-count-m4}

\subsection{Results on Large MoE Model}\label{sec:m4-bleu}

In Table 9
we provide aggregate BLEU scores for the results in Figure \ref{fig:m4-res}.

\begin{sidewaystable}[]

\resizebox{\linewidth}{!}{
\begin{tabular}{c|c|c|c|c|c|c|c|c|c|c|c}
\toprule
\multirow{2}{*}{{\bf System}} & \multicolumn{2}{c|}{{\bf Routing Granularity}} &\multicolumn{9}{c}{{\bf BLEU}}  \\
\cline{2-12}
& Encoder & Decoder & AVG & En-X & X-En & High-25 (EnX) & Mid 52 (EnX) & Low 25 (Enx) & High-25 (XEn) & Mid 52 (XEn) & Low 25 (XEn)  \\
\hline
Multilingual Transformer-Big   &    -   &     -   &  24.49  &  18.61 & 30.37   & 28.03  &  16.9  &  12.75 & 33.84 & 30.23 & 26.96 \\
Token-level MoE -- 128 experts     &   Token    &     Token   & \textbf{28.37}  & 20.51  & \textbf{36.26} & 30.99  & 18.94  & 13.33 & 40.14 & 36.74 & 31.03 \\

Task-level MoE -- 128 experts      &   Token    & Language & 28.09  & 20.66  & 35.52  & 31.21  & 19.17   & 13.28 & 39.69 & 36.42 & 29.16  \\
Task-level MoE -- 128 experts      &   Token    & Target & 27.83  & \textbf{20.76} & 34.90 & 31.05  & 19.23   & 13.68 & 38.88 & 35.28 & 29.93   \\

\bottomrule
\end{tabular}
}
\caption{We summarize the results in Figure \ref{fig:m4-res} on scaled up 128 expert MoE models. Here, \textit{High-25} means the average BLEU of the 25 highest resource languages, \textit{Low-25} means the average BLEU of the 25 lowest resource languages while \textit{Mid-52} is the average BLEU of the remaining 52 languages.}.
\end{sidewaystable}\label{tab:avg-bleu-m4}

\subsection{Gating Decisions for task-level and token-level MoEs}\label{sec:moe-analysis-sup}

In this section, we show the top expert distributions of different layers of the position-wise MoE model discussed in Section \ref{sec:moe-analysis} in Figures 6, 7, 8 and 9.

We also show expert distributions on MoE model routing by target language from EnX that was introduced in Section \ref{sec:m4-res2} in Figures 10 and 11.
We omit results on XEn language pairs because they belong to the same task in the context of this model.

\begin{figure*}

\begin{subfigure}{\textwidth}
    \centering
    \includegraphics[width=\textwidth]{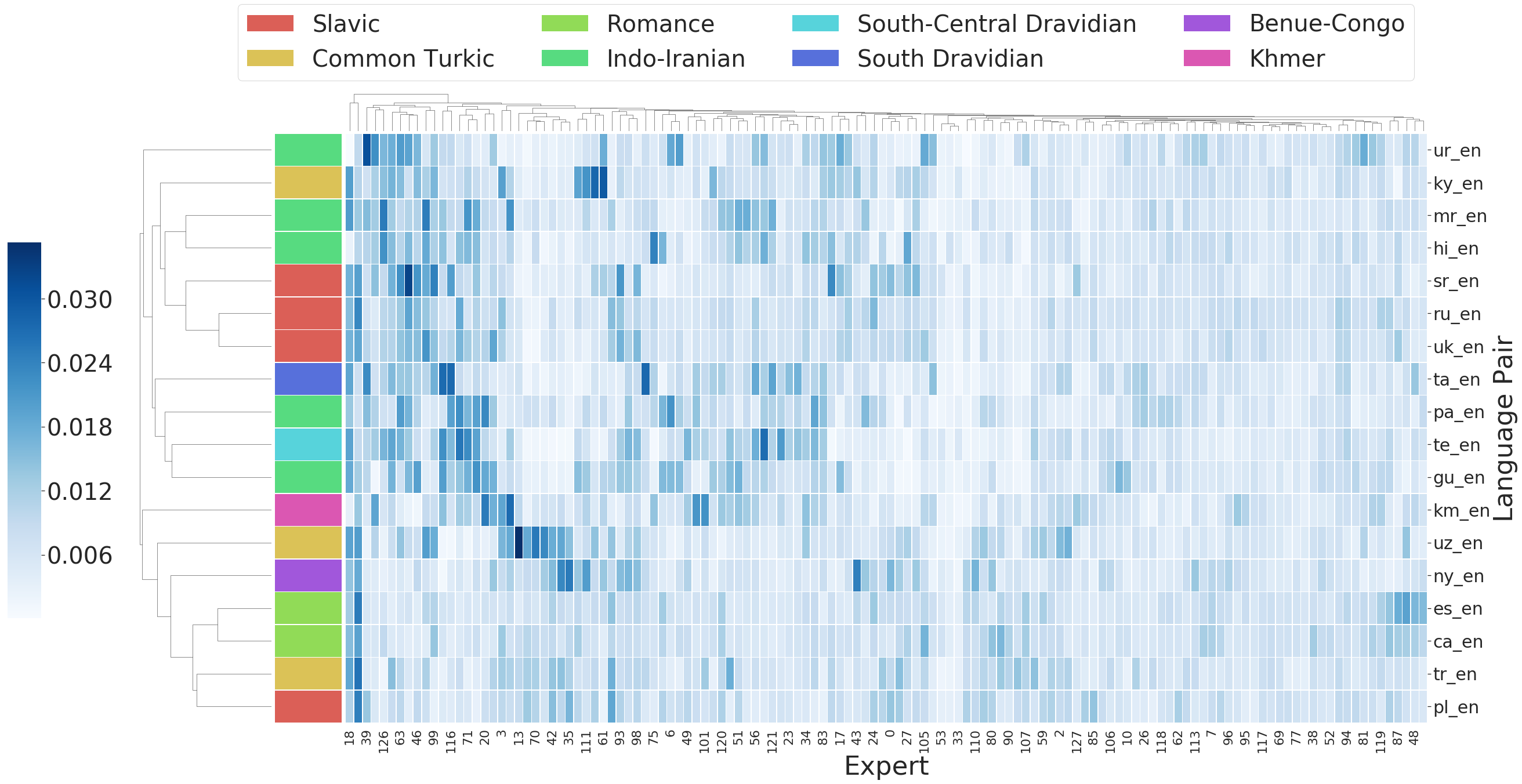}
    \caption{Gating decisions of the first layer of the encoder for Xx-En language pairs.}
    \label{fig:my_label}
\end{subfigure}

\begin{subfigure}{\textwidth}
    \centering
    \includegraphics[width=\textwidth]{figures/enc_layer6_xen.png}
    \caption{Gating decisions of the last layer of the encoder for Xx-En language pairs.}
    \label{fig:my_label}

\end{subfigure}

\caption{Gating decisions of the encoder of the position-wise MoE model on Xx-En language pairs, trained on internal data on a multiway parallel dataset. In this diagram, the darker a cell, corresponding to, say en-sr and the 37th expert, the more the expert is used. In both the last layer of the encoder and decoder, the tokens from each language are fairly well distributed across experts. In (a) the first layer of the encoder, there does not seem to be any major pattern in the expert distribution whereas in (b) the last layer of the encoder, tokens from all tasks (\textit{Xx-En}) seem to prefer the same set of few experts slightly over the others. }
\end{figure*}\label{fig:m4-xen-enc-sup}

\begin{figure*}

\begin{subfigure}{\textwidth}
    \centering
    \includegraphics[width=\textwidth]{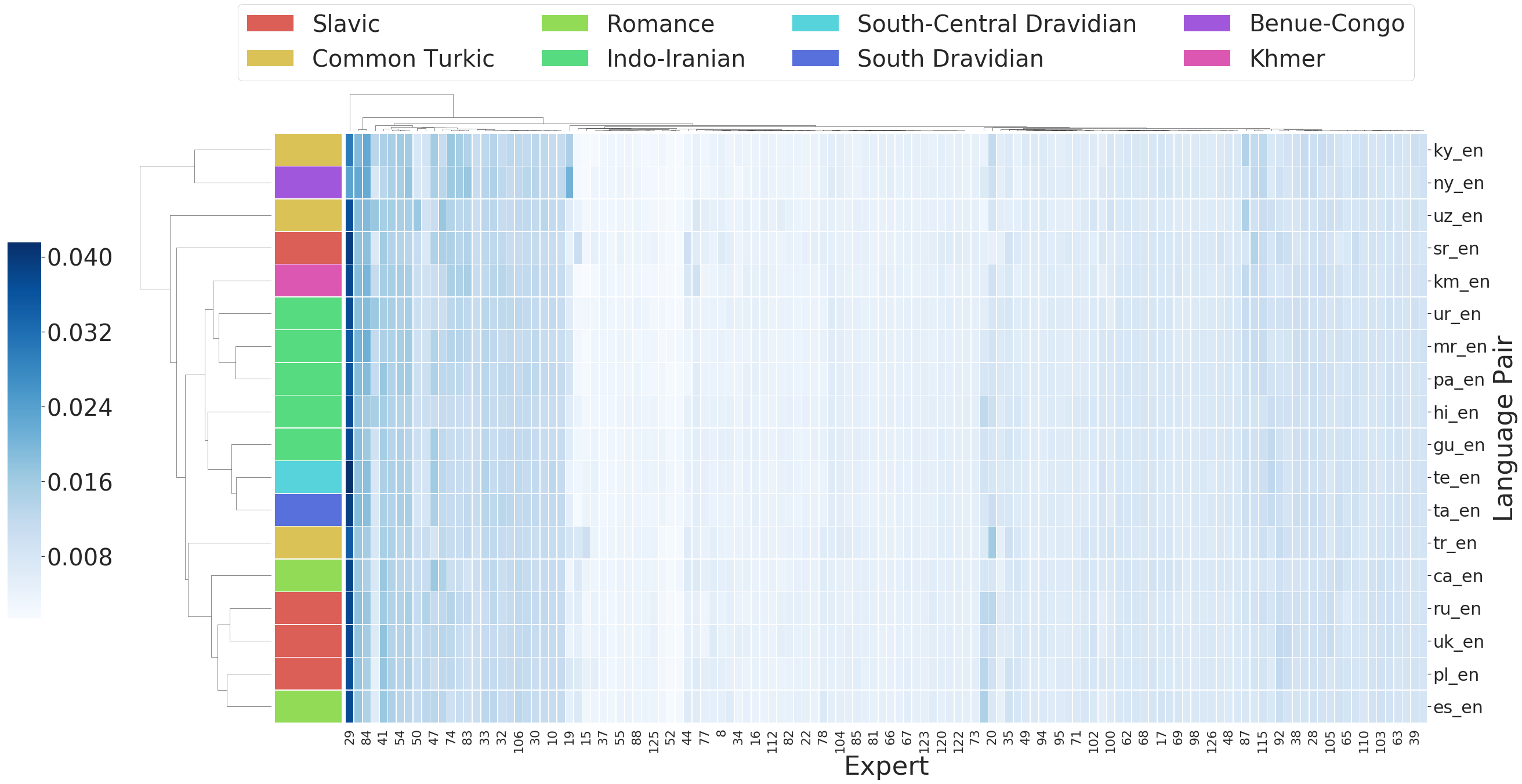}
    \caption{Gating decisions of the first layer of the decoder for Xx-En language pairs.}
    \label{fig:my_label}
\end{subfigure}

\begin{subfigure}{\textwidth}
    \centering
    \includegraphics[width=\textwidth]{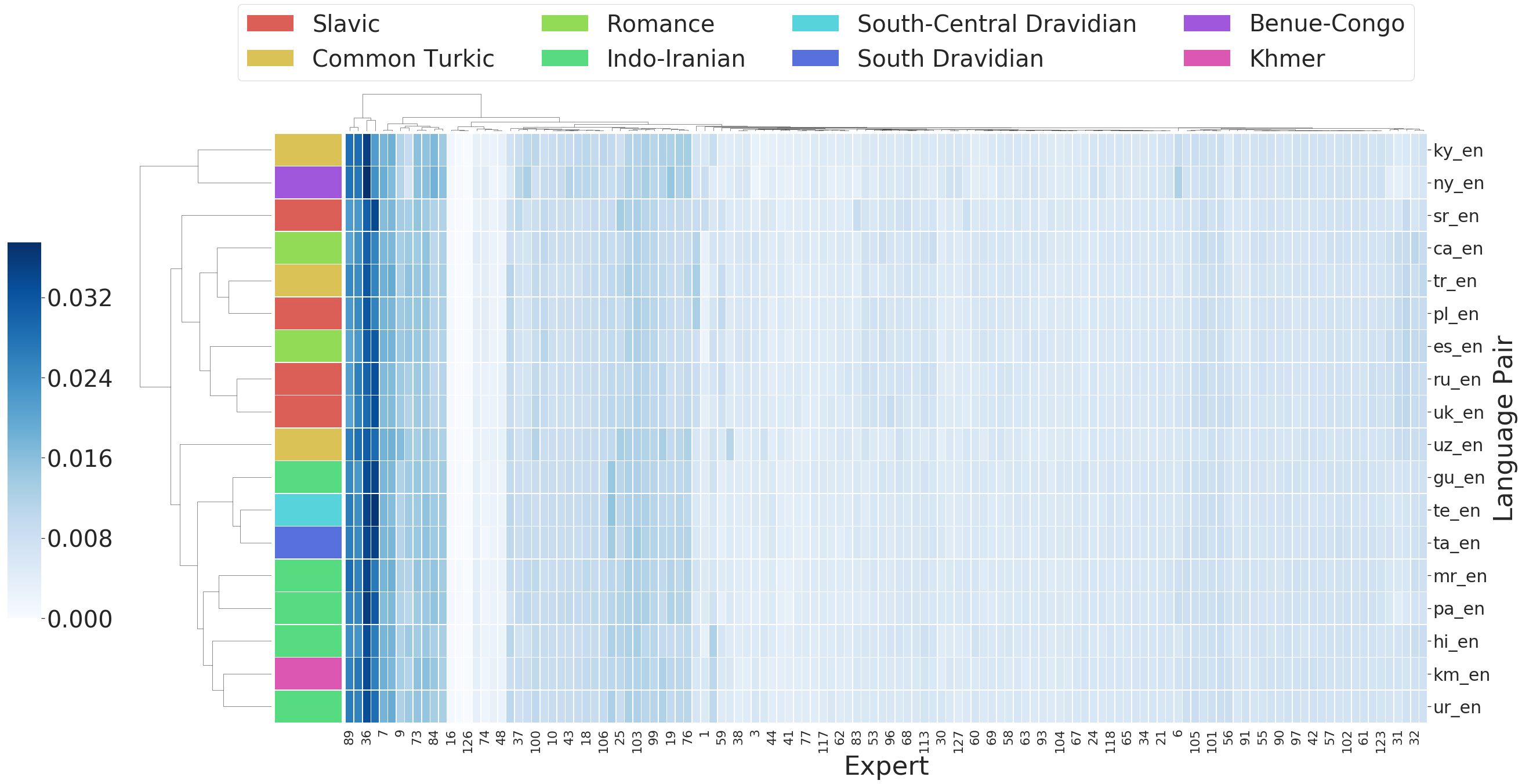}
    \caption{Gating decisions of the last layer of the decoder for Xx-En language pairs.}
    \label{fig:my_label}
\end{subfigure}
\caption{Gating decisions of the decoder of the position-wise MoE model on Xx-En language pairs, trained on internal data on a multiway parallel dataset. In this diagram, the darker a cell, corresponding to, say en-sr and the 37th expert, the more the expert is used. In both the first and last layer of the decoder, the tokens from each language are fairly well distributed across experts. In fact, tokens from all tasks (\textit{Xx-En}) seem to prefer the same set of few experts slightly over the others.}
\end{figure*}\label{fig:m4-xen-dec-sup}

\begin{figure*}

\begin{subfigure}{\textwidth}
    \centering
    \includegraphics[width=\textwidth]{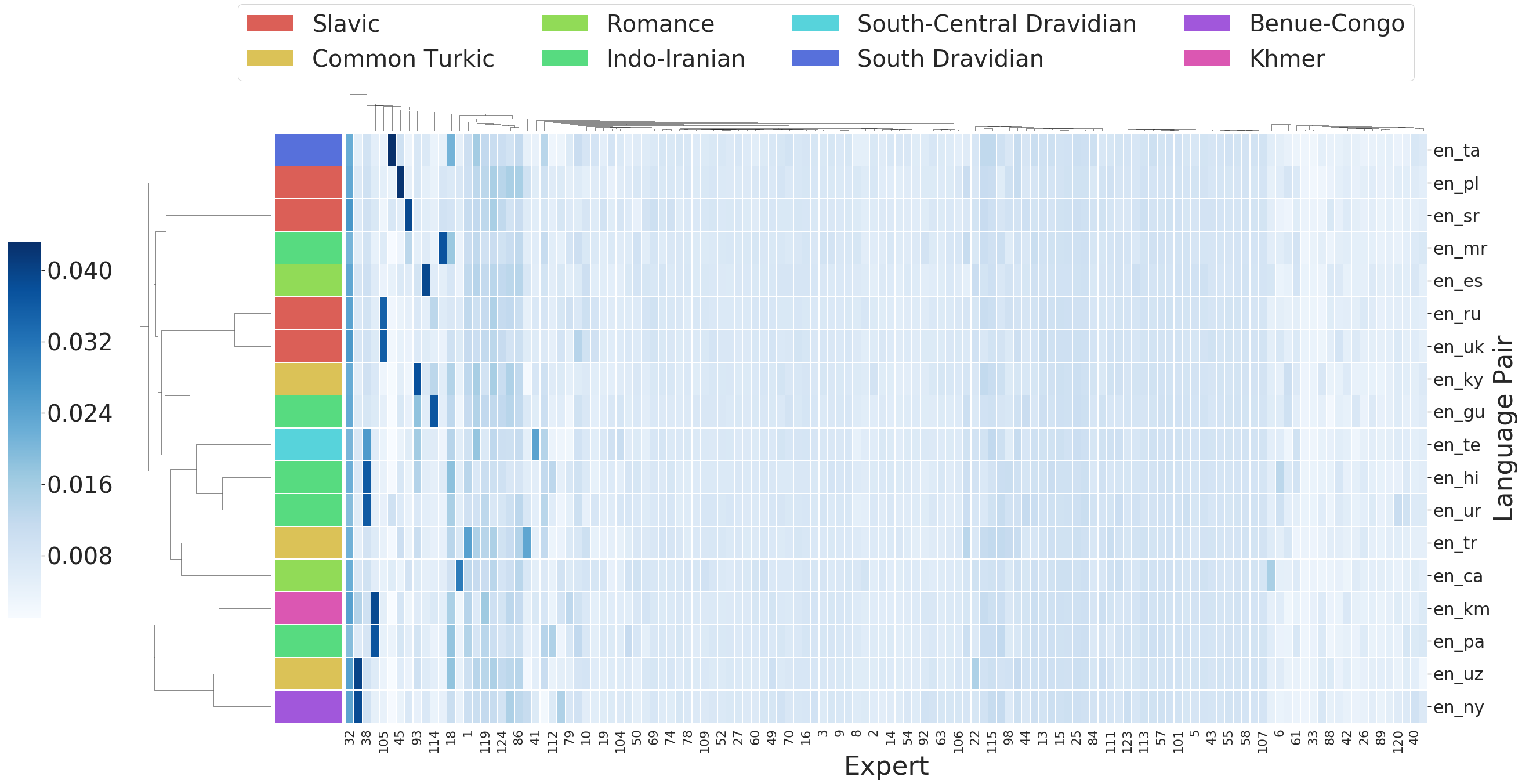}
    \caption{Gating decisions of the first layer of the encoder for En-Xx language pairs.}
    \label{fig:my_label}
\end{subfigure}

\begin{subfigure}{\textwidth}
    \centering
    \includegraphics[width=\textwidth]{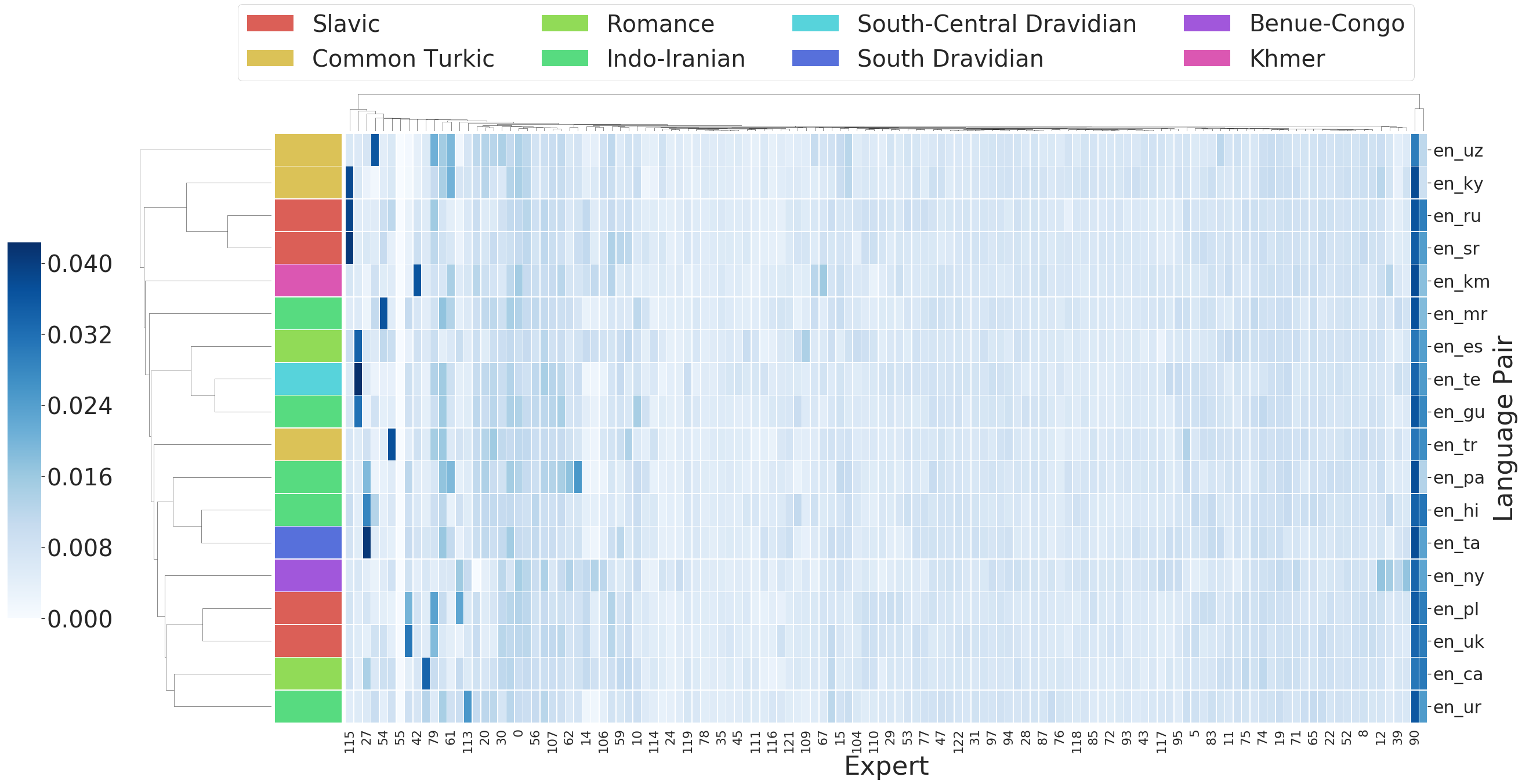}
    \caption{Gating decisions of the last layer of the encoder for En-Xx language pairs.}
    \label{fig:my_label}

\end{subfigure}
\caption{Gating decisions of the encoder of the position-wise MoE model on En-Xx language pairs, trained on internal data on a multiway parallel dataset. In this diagram, the darker a cell, corresponding to, say en-sr and the 37th expert, the more the expert is used. In both the first and last layer of the encoder, the tokens from each language are fairly well distributed across experts. Each task (\textit{En-Xx}) seems to slightly prefer a few experts over the other. }
\end{figure*}\label{fig:m4-enx-enc-sup}

\begin{figure*}

\begin{subfigure}{\textwidth}
    \centering
    \includegraphics[width=\textwidth]{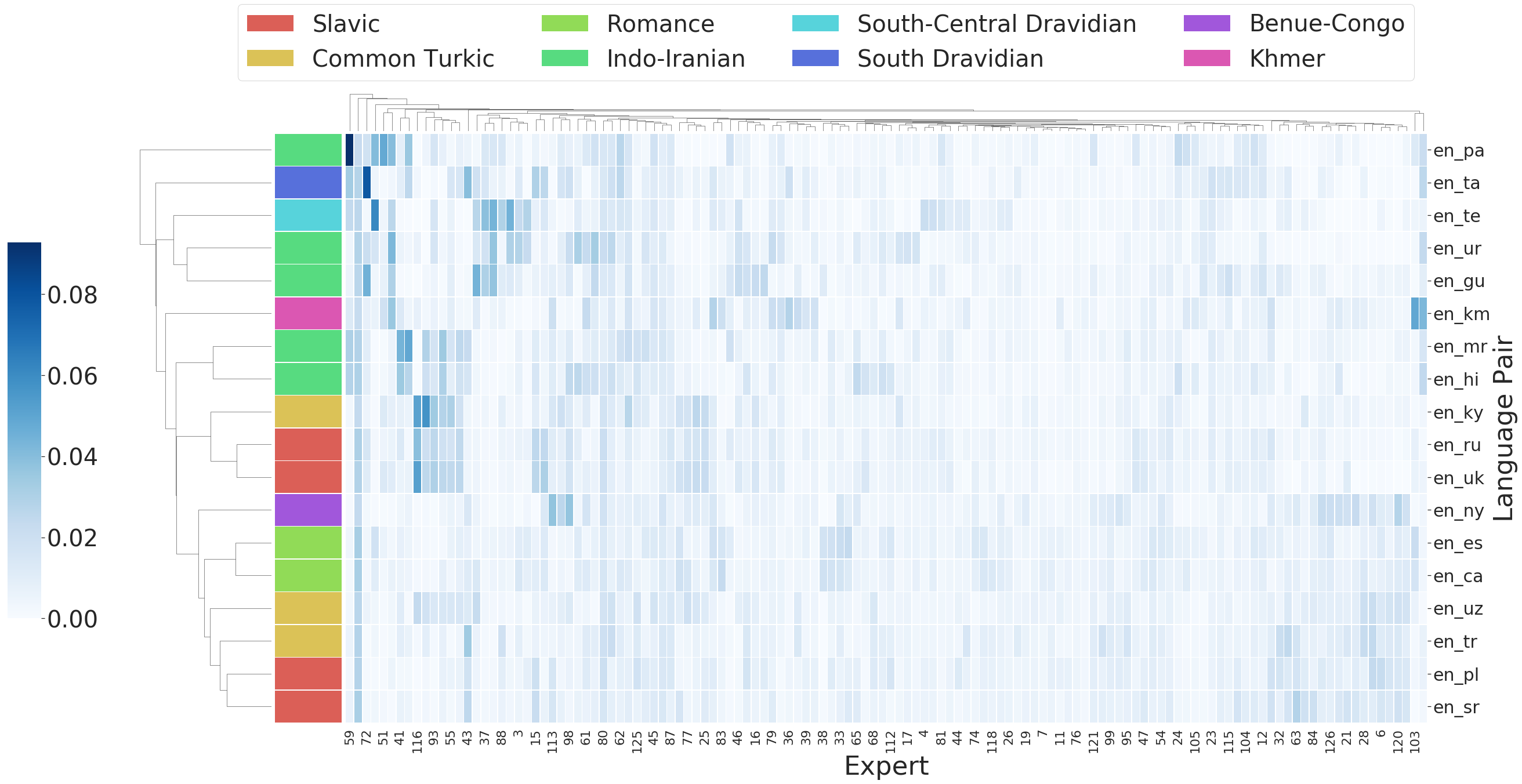}
    \caption{Gating decisions of the first layer of the decoder for En-Xx language pairs.}
    \label{fig:my_label}
\end{subfigure}

\begin{subfigure}{\textwidth}
    \centering
    \includegraphics[width=\textwidth]{figures/dec_layer6_enx.png}
    \caption{Gating decisions of the last layer of the decoder for En-Xx language pairs.}
    \label{fig:my_label}
\end{subfigure}
\caption{Gating decisions of the decoder of the position-wise MoE model on En-Xx language pairs, trained on internal data on a multiway parallel dataset. In this diagram, the darker a cell, corresponding to, say en-sr and the 37th expert, the more the expert is used. In both the first and last layer of the decoder, the tokens from each language are fairly well distributed across experts. Each task (\textit{En-Xx}) seems to slightly prefer a few experts over the other. Moreover, the set of experts appears to be similar for related languages.  For example, English-Spanish and English-Catalan (two Romance Languages) have similar expert distributions and so do English-Russian and English-Ukranian (two Slavic Languages).}
\end{figure*}\label{fig:m4-enx-dec-sup}
\begin{figure*}

\begin{subfigure}{\textwidth}
    \centering
    \includegraphics[width=\textwidth]{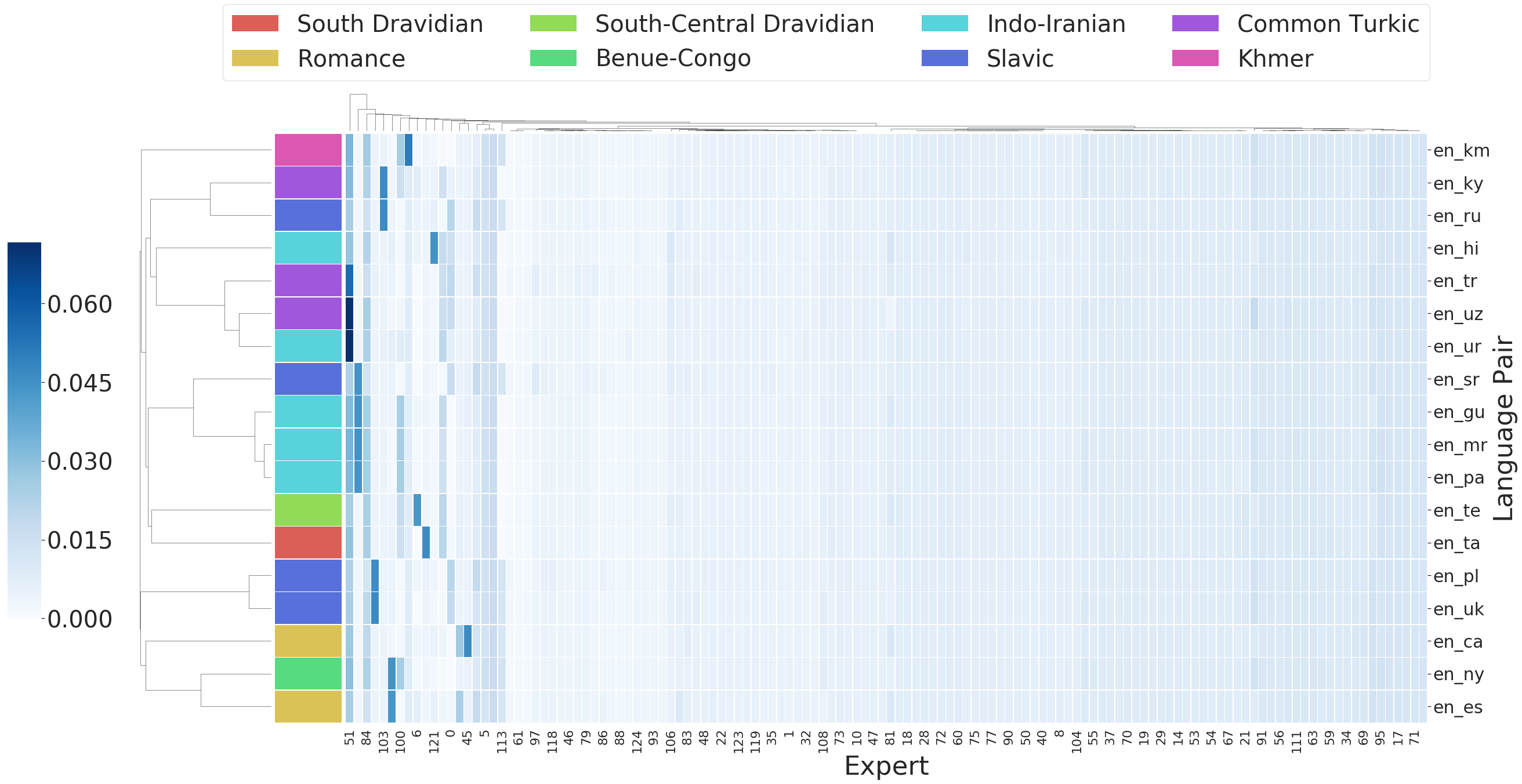}
    \caption{Gating decisions of the first layer of the encoder for En-Xx language pairs.}
    \label{fig:my_label}
\end{subfigure}

\begin{subfigure}{\textwidth}
    \centering
    \includegraphics[width=\textwidth]{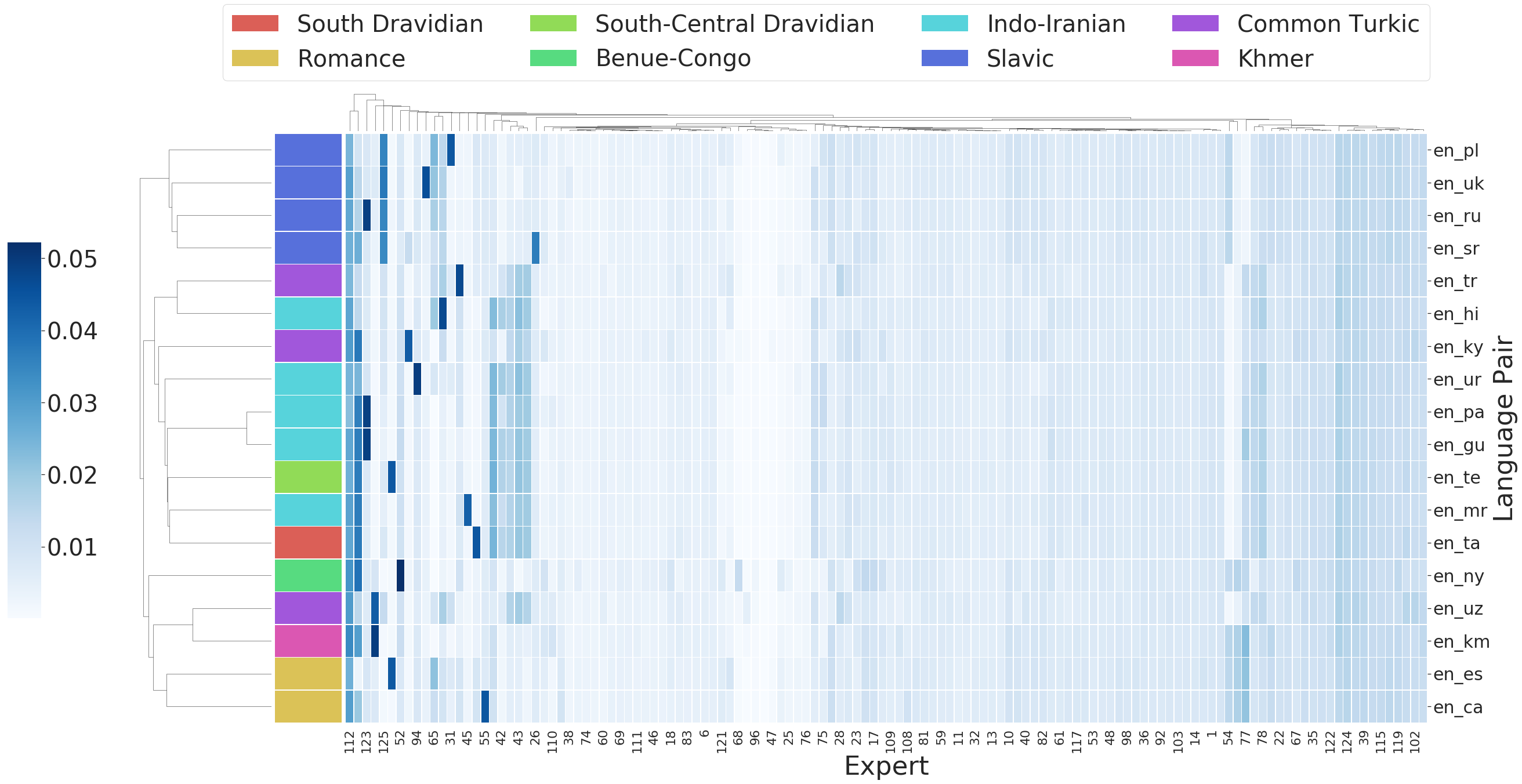}
    \caption{Gating decisions of the last layer of the encoder for En-Xx language pairs.}
    \label{fig:my_label}

\end{subfigure}
\caption{Gating decisions of the encoder of the target language-wise MoE model on En-Xx language pairs, trained on internal data on a multiway parallel dataset. In this diagram, the darker a cell, corresponding to, say en-sr and the 37th expert, the more the expert is used. The encoder behaves similarly to that of the position-wise model: in both the first and last layer of the encoder, the tokens from each language are fairly well distributed across experts. Each task (\textit{En-Xx}) seems to slightly prefer a few experts over the other. }
\end{figure*}\label{fig:task-m4-enx-enc-sup}

\begin{figure*}

\begin{subfigure}{\textwidth}
    \centering
    \includegraphics[width=\textwidth]{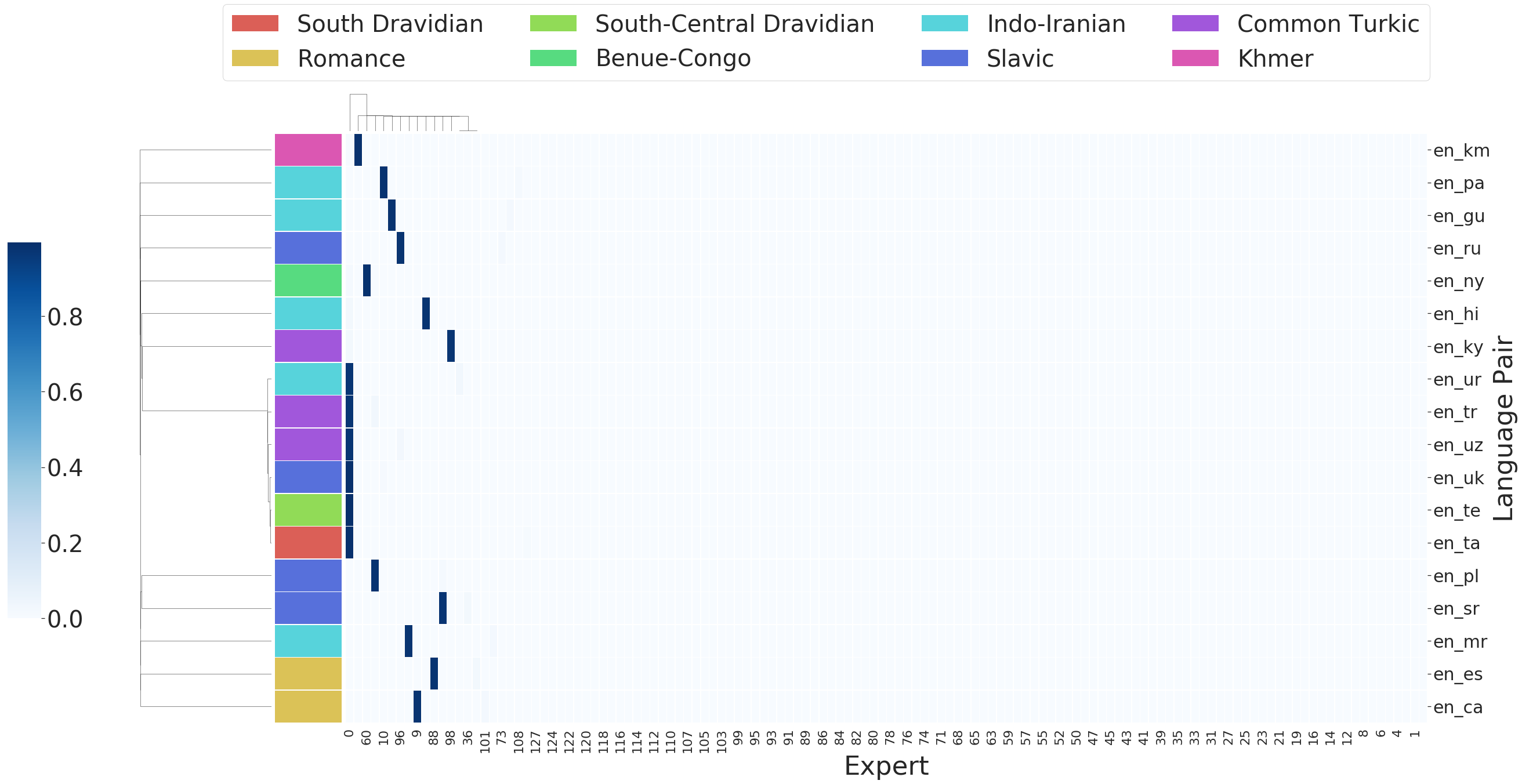}
    \caption{Gating decisions of the first layer of the decoder for En-Xx language pairs.}
    \label{fig:my_label}
\end{subfigure}

\begin{subfigure}{\textwidth}
    \centering
    \includegraphics[width=\textwidth]{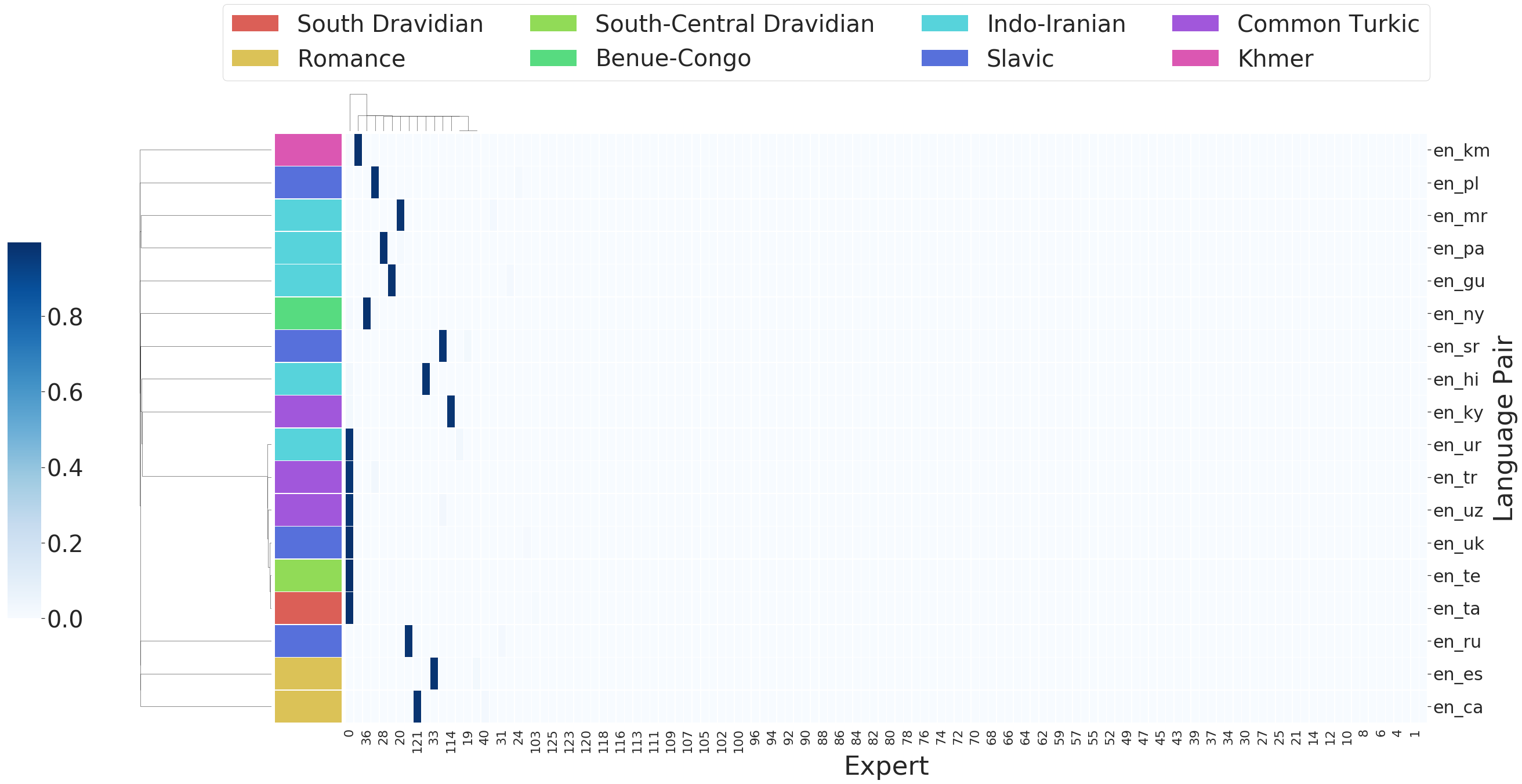}
    \caption{Gating decisions of the last layer of the decoder for En-Xx language pairs.}
    \label{fig:my_label}
\end{subfigure}
\caption{Gating decisions of the decoder of the target language-wise MoE model on En-Xx language pairs, trained on internal data on a multiway parallel dataset. In this diagram, the darker a cell, corresponding to, say en-sr and the 37th expert, the more the expert is used. There seems to be some amount of expert sharing on a linguistic basis: en-ur, en-te and en-ta (two Dravidian Languages and an Indo-Iranian language) and en-tr, en-uz and en-uk (two Turkic languages and a Slavic language) share an expert. On the other hand, en-es and en-ca (two Romance languages) have different experts. }
\end{figure*}\label{fig:task-m4-enx-dec-sup}

\end{document}